\pdfoutput=1

\documentclass[10pt,journal,compsoc]{IEEEtran}

\usepackage{amsmath,amsfonts,bm}

\def\eqref#1{equation~\ref{#1}}

\def\1{\bm{1}}

\DeclareMathAlphabet{\mathsfit}{\encodingdefault}{\sfdefault}{m}{sl}
\SetMathAlphabet{\mathsfit}{bold}{\encodingdefault}{\sfdefault}{bx}{n}

\DeclareMathOperator*{\argmax}{arg\,max}

\ifCLASSOPTIONcompsoc
  \usepackage[nocompress]{cite}
\else
  \usepackage{cite}
\fi

\ifCLASSINFOpdf
\else
\fi

\usepackage[x11names,usenames,dvipsnames]{xcolor}
\usepackage{forest}
\usepackage{bm}
\usepackage{array}
\usepackage{amssymb}
\usepackage{amsmath}
\usepackage{subfiles}
\usepackage{ragged2e}
\usepackage{multirow}
\usepackage{longtable}
\usepackage{graphicx}
\usepackage{subfigure}
\usepackage{bbding}
\usepackage{booktabs} 
\usepackage{bbm}
\usepackage{float}
\usepackage[square, comma, sort&compress, numbers]{natbib}
\usepackage{tabularx}
\usepackage{tikz}
\usepackage{graphics}
\usepackage{pgfplots}
\pgfplotsset{compat=1.17}
\usepackage{newtxmath}

\usepackage{graphicx}

\usepackage{algorithm}

\usepackage{algorithmic}
\usepackage{tabularx}
\usepackage[backref]{hyperref}
\hypersetup{
    colorlinks=true, 
    linkcolor=black, 
    citecolor=black, 
    filecolor=black, 
    urlcolor=black, 
    pdfnewwindow=false, 
    linktoc=none 
}
\usepackage{etoolbox}
\AtBeginEnvironment{IEEEbiography}{\vspace{-2.6\baselineskip}}
\usepackage{float}
\usepackage{multirow}
\usepackage{url}
\usepackage[numbers,sort&compress]{natbib}
\newtheorem{Theorem}{Theorem}

\begin{document}

\title{An Extensive Survey with Empirical Studies on \\ Deep Temporal Point Process}

\author{~Haitao~Lin$^{\dag, 1,2}$, ~Cheng~Tan$^{1,2}$,~Lirong~Wu$^{1,2}$,\\ ~Zicheng~Liu$^{1,2}$ ~Zhangyang~Gao$^{1,2}$, and~Stan.Z.Li$^{*,2}$
        
\IEEEcompsocitemizethanks{\IEEEcompsocthanksitem $\dag$ First Author. Email: linhaitao@westlake.edu.cn. 
\IEEEcompsocthanksitem 1. Zhejiang University, Hangzhou, China.
\IEEEcompsocthanksitem 2. Center for Artificial Intelligence Research and Innovation, School of Engineering, Westlake University, Hangzhou, China
\IEEEcompsocthanksitem Corresponding Author. IEEE Fellow. Email: Stan.ZQ.Li@westlake.edu.cn. 
}}

\markboth{Journal of \LaTeX\ Class Files,}%
{Shell \MakeLowercase{\textit{\textit{et al.}}}: Bare Demo of IEEEtran.cls for Computer Society Journals}

\IEEEtitleabstractindextext{
\begin{abstract}
Temporal point process as the stochastic process on a continuous domain of time is commonly used to model the asynchronous event sequence featuring occurrence timestamps.
Thanks to the strong expressivity of deep neural networks, they are emerging as a promising choice for capturing the patterns in asynchronous sequences, in the context of temporal point process.
In this paper,  we first review recent research emphasis and difficulties in modeling asynchronous event sequences with deep temporal point process, which can be concluded into four fields: encoding of history sequence, formulation of conditional intensity function, relational discovery of events, and learning approaches for optimization. 
We introduce most of the recently proposed models by dismantling them into four parts and conduct experiments by re-modularizing the first three parts with the same learning strategy for a fair empirical evaluation.
Besides, we extend the history encoders and conditional intensity function family and propose a Granger causality discovery framework for exploiting the relations among multi-types of events.
Because the Granger causality can be represented by the Granger causality graph, discrete graph structure learning in the framework of Variational Inference is employed to reveal latent structures of the graph. 
Further experiments show that the proposed framework with latent graph discovery can both capture the relations and achieve an improved fitting and predicting performance.
\end{abstract}

}

\maketitle
\IEEEdisplaynontitleabstractindextext
\IEEEpeerreviewmaketitle

\section{Introduction}\label{sec:introduction}
Asynchronous event sequences are generated ubiquitously by human behaviors or natural phenomena, such as electronic patients' records, financial transactions, extreme geophysical event occurrence, and so on. Studying the temporal distribution of events and discovering the relationships among different types of events is of great scientific interest for understanding the dynamics and mechanism of the occurrence of the events. 
One of the choices for it is temporal point process \cite{pointprocess}, defined as the stochastic processes with marked events on the continuous domain of time, which can naturally capture the clustering \citep{hawkesprocess} or self-correcting \citep{ISHAM1979335} phenomena of such sequences of events. Usually, modeling the rate of event occurrence known as conditional intensity, as a function of time given the previous observation of events, is a solution to capturing the dynamics of the process. 
Since the distribution of such a process is completely governed by the conditional intensity function, statistical prediction and inference can all be conducted via the conditional intensity functions.

Although a series of works have achieved remarkable progress in temporal point process, particularly in deep-neural-network-based models \citep{sahp,adversarial,rmtpp,lognorm,fnn,ctlstm}, the majority of them use different history encoders to embed history events, and different forms of intensity functions parameterized by the embedded historical sequence of events. This casts a problem to us: \textit{Which part is paramount to the performance improvements?} 
Thus we firstly review most methods on the current deep temporal point process, and then dismantle them into four major parts: encoding of history sequence, formulation of conditional intensity function, relational discovery of events and learning approaches for optimization. Then we extend and reassemble the first three parts except the learning approaches in experiments, for a fair empirical evaluation as well as deep insights into each part.

Aside from modeling the dynamics and improving predictive performance, we find that the studies focusing on discovering the latent relation of different types of events are still rare, especially in fields of deep neural networks, which are usually considered as lacking in interpretability. 
The challenges in relational inference can be viewed as a graph structure learning task, and recently, relevant techniques in Graph Neural Networks have been proposed as a solution to it. Moreover, in classical statistical learning, Granger causality \citep{graphicalhawkes, icmlgranger} which can be represented by Granger causality graphs, provides a more interpretable and stronger definition of relations among multivariate time series and draws more research attention. 
Since few work on this field has been proposed although it is a key to interpretability of the process, we aims to propose a framework for inter-events Granger causality discovery, which can be flexibly applicable to the existing methods.

In summary, the outline of the paper can be listed as:
\begin{figure*}[htb]
    \centering
    \includegraphics[width=7.6in]{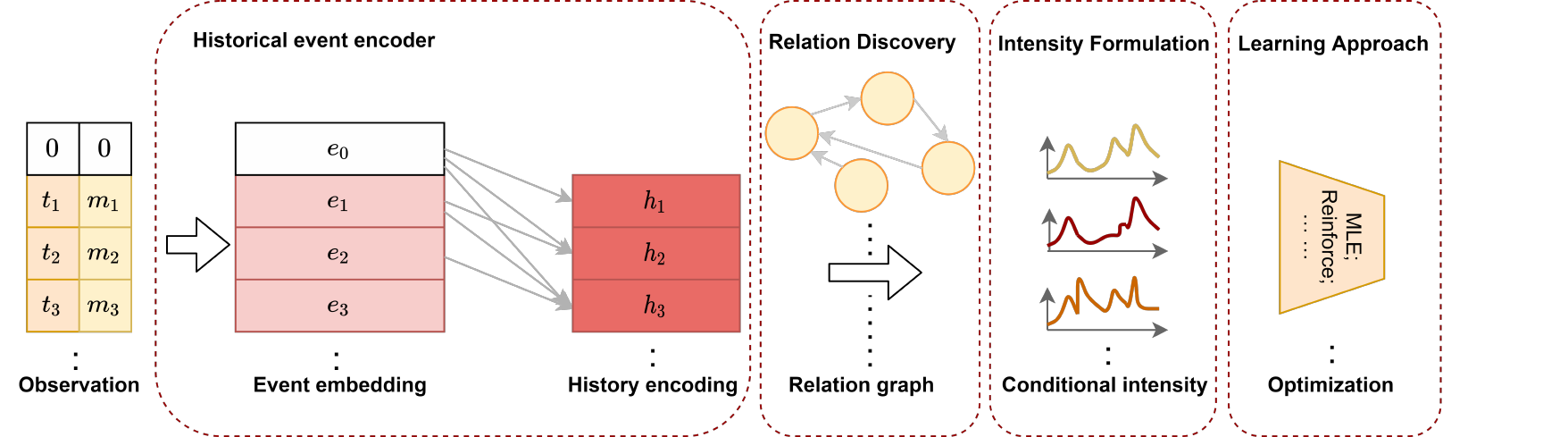}
    \caption{The workflows of deep temporal point process are divided into the four parts: encoding of history sequence,  relational discovery of events,  formulation of conditional intensity function and learning approaches.}
    \label{figure:fourparts}
\end{figure*}\vspace*{-0.3cm}
\begin{itemize}
    \item[\textbf{Sec.2}] Conclusion of the four parts -- Dividing the existing methods of deep temporal point process into the four parts: encoding of history sequence, formulation of conditional intensity function,  relational discovery of events and learning approaches, and giving a united formulation on them.
    \item[\textbf{Sec.3}] Modification and extension of the four parts -- Revising the four parts which are adopted by most existing methods, and extending some of the parts, including the history encoders with a modified FNet \citep{Fnet} and the conditional intensity functions with a family of mixture distributions supported on a semi-infinite interval.
    \item[\textbf{Sec.4}] Review of the existing methods according to the four parts -- Carrying out 'case study', i.e. dismantling most of existing methods into the four parts to improve the clarity of each method for further fair empirical evaluation. 
    \item[\textbf{Sec.5}] Development of a Granger causality discovery framework in deep temporal point process.
    \item[\textbf{Sec.6}] Demonstration of model performance through fair empirical study for deep temporal point process: 
\end{itemize}
\begin{itemize}
    \item Conducting experiments of the different combinations of the first two parts to evaluate which part is of most significant importance on performance;
    \item Conducting experiments of the proposed framework, to show that it can reveal the Granger causalities as well as achieve competitive performance on real-world datasets.
\end{itemize} 
\begin{itemize}
    \item[\textbf{Sec.7}] Discussion on existing problems and promising directions for future works.
\end{itemize}

In this way, the topic of our work is Extensive Deep Temporal Point Process (EDTPP), and we aim to integrate existing models, facilitate further expansions and explore the existing problems for future research. Our source code of EDTPP is available on \url{https://github.com/BIRD-TAO/EDTPP}.
\section{Review of Deep Temporal Point Process}
\subsection{Premilinaries of temporal point process}
In marked temporal point process, the observation is represented by a sequence of event time stamps $\{t_i\}_{1 \leq i\leq N}$ and markers $\{m_i\}_{1 \leq i\leq N}$, such that $t_i\in [0,T)$, $t_i < t_{i+1}$, $\forall i \geq 1$. A set of right-continuous counting measures $\{\mathcal{N}_m(t)\}_{m\leq M}$, the measure in which is defined as the number of events occurring in the time interval $[0,t)$ of the type-$m$ event, where there are $M$ types of events and $m \in [M]$ with $[M] =\{1,2,\ldots,M\}$. 
Given the history $\mathcal{H}(t) = \{(t_j, m_j), t_j < t\}$, the temporal point process can be characterized via its conditional intensity function (CIF), defined as 
\begin{align}
\label{eq:intensitydefine}
    \lambda_m^*(t) =& \quad\lambda_m (t|\mathcal{H}(t)) \\
    =& \lim_{\Delta t \rightarrow 0^+} \frac{\mathbb{E}[\mathcal{N}_m(t + \Delta t) - \mathcal{N}_m(t)| \mathcal{H}(t)]}{\Delta t}, \notag
\end{align}
which means the expected instantaneous rate of happening the events given the history. Note that it is always a non-negative function of $t$. Given the CIF, the probability density function (PDF) of the type-$m$ event reads
\begin{align}
    f^*_m(t) =  \lambda_m^*(t)\exp(-\int_{t_{i-1}}^t \lambda_m^*(\tau)d\tau), \label{eq:pdf}
\end{align}
where $i-1 = \argmax_{j\leq n}\{t_j, t_j < t\}$. 
The leading target of deep temporal point process is to parameterize a model to fit the distribution of the generated marked timestamps, as to inference PDF or CIF for further statistical prediction, including next event time and type prediction.
More details on preliminaries are given in Appendix A.
\subsection{Historical event encoder}
Either CIF or PDF is a function of both $t$ and historical events before $t$, i.e. $\mathcal{H}(t)$.
Therefore, to model the process, the first issue to solve is \textbf{how to embed the history sequence of events for formulating the CIF or PDF} of the occurrence of different types of events. For the $i$-th event's history $\mathcal{H}(t_i)$, $j$-th event in the history set is embedded in a high-dimensional space, considering both temporal and type information, as
\begin{align}
    \bm {e}_j = [\bm{\omega}(t_j); \bm{E}^T\bm{m}_j], \label{eq:eventemb}
\end{align}
$\bm{\omega}$ transform one-dimension $t_j$ into a high-dimension vector, which can be linear, trigonometric, and so on, $\bm{E}$ is the embedding matrix for event types, and $\bm{m}_j$ is the one-hot encoding of event type $m_j$.
A historical encoder $\bm{H}$ thus maps the sequence of embedding $\{\bm{e}_1,\bm{e}_2,\ldots,\bm{e}_{i-1}\}$ into a vector space of dimension $D$, by
\begin{align}
    \bm{h}_i = \bm{H}([\bm{e}_1;\bm{e}_2;\ldots;\bm{e}_{i-1}]).\label{eq:histenc}
\end{align}
$\bm{H}$ can be chosen as recurrent sequence encoders, and $\bm{h}_i \in \mathbb{R}^D$ will be used for the parameterization of the CIF. 

\subsection{Conditional intensity formulation}
The conditional intensity function with parameters $\bm{\Theta}_{m}(t)$ is written as  $\lambda_{m}(t; \bm{\Theta}_{m}(t)|\mathcal{H}(t))$. The parameter $\bm{\Theta}_{m}(t)$ is assumed as a piece-wise function of $t$, i.e.
\begin{align}
\label{eq:intensitytrans}
    \bm{\Theta}_m(t) = \bm{\chi}_m(\bm{h}_i) \quad\quad t\in[t_{i-1}, t_{i}).
\end{align}
From another view, the new occurrence of the $m$-type event will make difference to $\bm{h}_i$, and thus update the $\bm{\Theta}_m(t)$. 

The expressivity of the chosen family of functions to approximate the target CIF is of great significance. The better is the approximating ability of the chosen family of CIF, the better fitting performance the model of deep temporal point process will achieve. 
Besides, as showed in Eq. \ref{eq:pdf}, to maximize the likelihood of the observed sequence of events, the integral term is unavoidable, so the difficulties in tackling the log-likelihood are the high computational cost from the calculation of the integral term, and the closed-form of this term leads the computation of likelihood manageable.
In conclusion, \textbf{how to approximate the target CIF of the process with a family of functions} is the second issue. The functions used to approximate the target should possess both closed integral form and powerful expressivity.

\subsection{Relation discovery of events}
\begin{figure}[!htb]
    \includegraphics[width=3.2in]{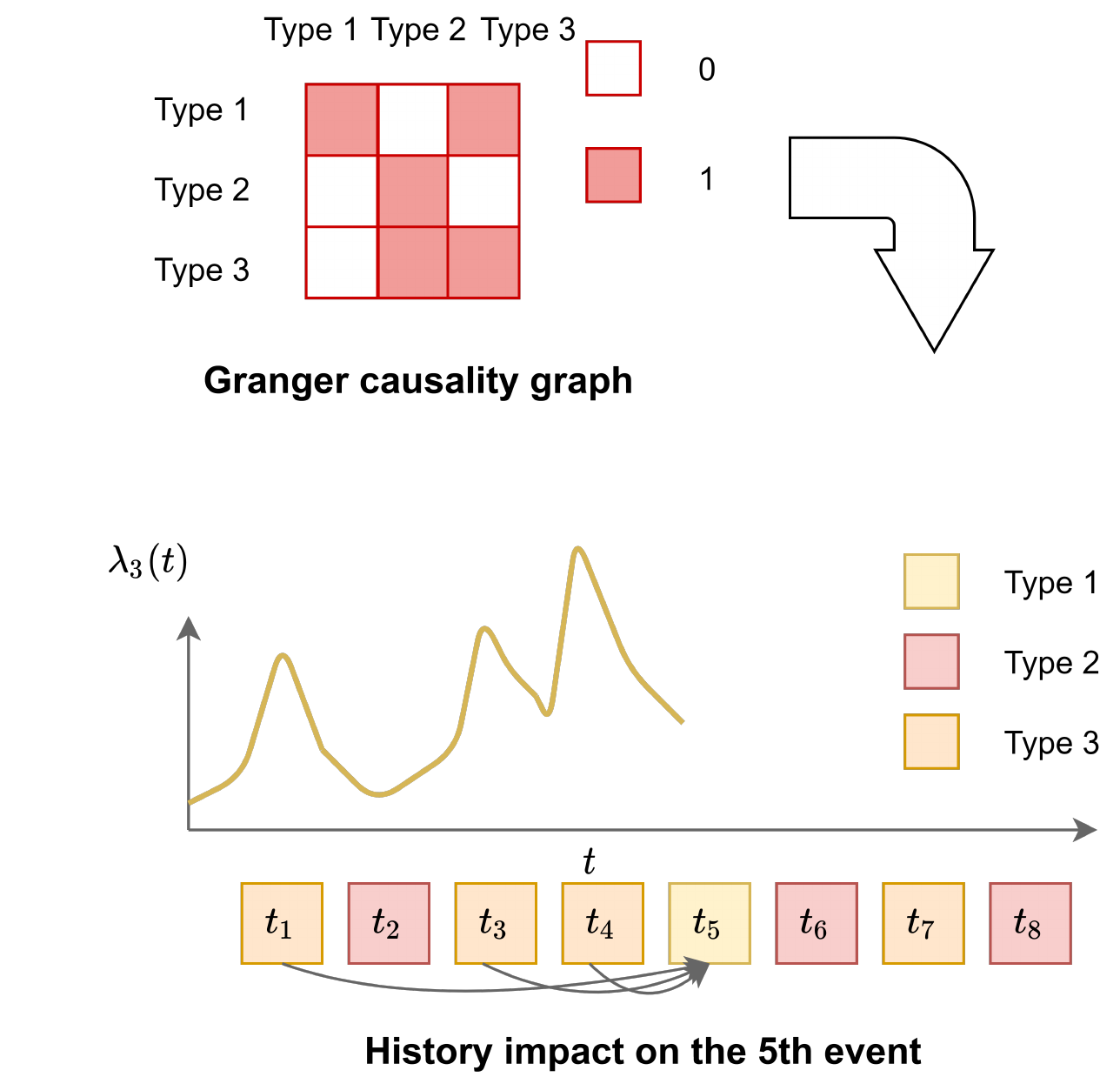}
    \caption{An example of Granger causality graph representing the Granger causality of events. Events of type 1 are affected by type 3 and itself according to the Granger causality graph, so the CIF of type 1 events before $t_5$ is augmented only when type 1 and 3 events happen. }
    \label{figure:grangercauseexample}
  \end{figure}
There are two kinds of methods for relational discovery in deep temporal point process -- model-agnostic \cite{zhang2020cause} and model-specific \cite{dgnpp}. We focus on the second kind which aims to propose an end-to-end framework for both discovering the relation as well as fitting the process. To inference the implicit relations among different types of events, graph structure learning is usually harnessed, in which event types are regarded as nodes, and the pairwise relations are considered as edges between types. 

Different from generalized relation discovery, in classical statistical learning on time series, a more interpretable and stronger relation known as Granger causality draws more interests, which measures whether a previous occurrence of a certain type of event will influence the occurrence of another type in the future. 
To proceed formally \citep{hawkesprocess}, for any $\mathcal{M} \subseteq [M]$, natural filtration expanded by the
sub-process $\{\mathcal{N}_{m}(t)\}_{m \in \mathcal{M}}$ is denoted by $\mathcal{H}_{\mathcal{M}}(t) = \{\mathcal{H}_{m}(t)\}_{m \in \mathcal{M}}$, which is the sequence of smallest $\sigma$-algebra expanded by the event history of type $m \in \mathcal{M}$, i.e., $\mathcal{H}_{\mathcal{M}}(t) = \bm{\Sigma}(\{\mathcal{N}_m(\tau),| m\in \mathcal{M}, \tau < t\})$. We further write $\mathcal{H}_{-m}(t) = \mathcal{H}_{[M]\setminus\{m\}}(t)$, for any $m\in[M]$.
\\
\textit{\textbf{Definition 1.}} {\rm \citep{graphicalhawkes}} For a $M$-types temporal point process, event type $m$ is \textbf{Granger non-causal} for event type $m'$, if $\lambda_{m'}^*(t)$ is $\mathcal{H}_{-m}(t)$-measurable for all $t$; Otherwise, event type $m$ \textbf{Granger causes} event type $m'$.

From another perspective, the above definition can be concluded that if the changes of historical events of type $m$ which is $\mathcal{H}_m(t)=\{(t_i,m_i)|t_i<t, m_i=m\}$, do not have further impacts on $\lambda^*_{m'}(t)$ at any time $t$,  then we can say that type $m$ is Granger non-causal for type $m'$. Otherwise, type $m$ Granger-causes type-$m'$ event. 

In this way, we claim that the third issue to state is \textbf{how to reveal the latent Granger causality among events}, as a relational discovery problem which can be formulated as learning the structure of the following defined Granger causality graph:
\\
\textit{\textbf{Definition 2.}} A Granger causality graph $\mathcal{G} = (\bm{V}, \bm{\mathcal{E}}, \bm{A})$ can be established to represent the Granger causality of one event to another, where $\bm{V} = [M]$ is the vertex set, for each $m \in \bm{V}$ denoting a type of event, $\bm{\mathcal{E}}$ is the edge set with $(m,m')\in \bm{\mathcal{E}}$ representing that type-$m$ events Granger cause type $m'$, and $\bm{A}$ is the corresponding adjacency matrix with $\bm{A}_{m',m}=1$ meaning $(m,m')\in \bm{\mathcal{E}}$, or $\bm{A}_{m',m}=0$. A reasonable assumption is that $\bm{A}_{m,m} = 1$, implying that there is always a self-causality.

Therefore, the discovery of Granger causality among events is equivalent to learning the latent Granger causality graph's structure \cite{graphicalhawkes}, attributing to their one-to-one relations. In this way, approaches in graph structure learning can be solutions to relational inference or more specifically, Granger causality discovery in deep temporal point process.

\subsection{Learning approaches}
The final issue drawing research interest is the learning approaches and strategies, i.e. \textbf{ how to set up and optimize the target function to get a better fitting and predictive performance of the model}.
In statistical inference, maximum likelihood is most commonly used to fit the model. For temporal point process with $M$ types of events, given a sequence $\{(t_i, m_i)\}_{1\leq i \leq N}$, the log-likelihood is given by 
\begin{align}
    l(\bm{\Theta}) =& \sum_{i=1}^{N}\log \lambda_{m_i}(t_i; \bm{\Theta}_{m_i}(t)|\mathcal{H}(t)) \\ &- \sum_{m=1}^{M}\int_0^{T} \lambda_m(t; \bm{\Theta}_m(t)|\mathcal{H}(t))dt. \label{loglikelihood} \notag
\end{align}
The negative log-likelihood is usually optimized by stochastic gradient descent (SGD) methods, such as Adam\cite{kingma2017adam}.
When the method is directly modeling the PDF $f^*_m(t)$, the target function can be written as 
\begin{align}
    l(\bm{\Theta}) =& \sum_{i=1}^{N}\log f_{m_i}(t_i; \bm{\Theta}_{m_i}(t)|\mathcal{H}(t))
\end{align}
\begin{figure*}
    \centering
    \includegraphics[width=7.2in]{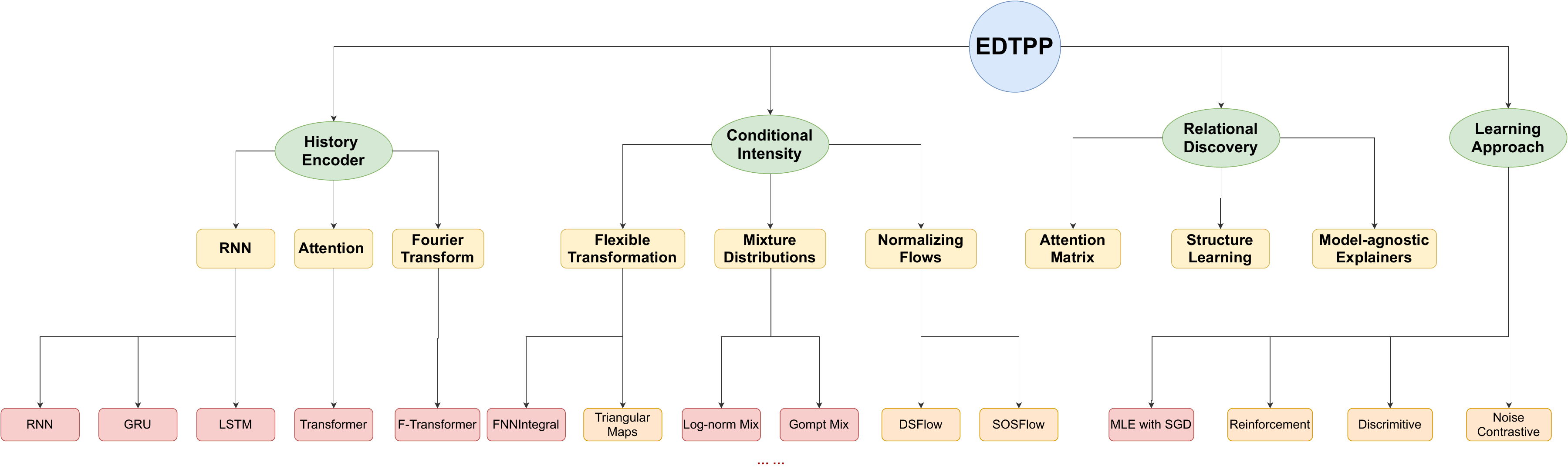}
    \caption{The dismantled four leading parts of EDTPP. In the final row, the box in red means that it is implemented in our code for fair empirical study, while the box in orange means that it will be further added for completeness.}
    \label{figure:hierarchygraph}
  \end{figure*}
Note that the two targets are equivalent, according to Eq.~\ref{eq:intensitydefine}.
There are a series of learning approaches, including reinforcement learning \citep{reinforcement1, reinforcement2}, adversarial learning \citep{adversarial}, noise contrastive learning \citep{noisecontrast}, which will be discussed later.

\subsection{Next event prediction}
Besides the task of fitting the distribution of the next event arrival time, when there exists more than one event type, another predictive target is what type of event is most likely to happen, given the historical observations. The task can be regarded as a categorical classification, and usually achieved through firstly transforming the history encoding $\bm{h}_{i-1}$ to  discrete distribution's logit scores as 
\begin{align}
    \kappa(\bm{h}_{i}) = \mathrm{logit}(\hat m_i), \label{eq:typelogit}
\end{align}
where $\mathrm{logit}(\hat m_i)\in \mathbb{R}^{M}$, $\kappa : \mathbb{R}^D \rightarrow \mathbb{R}^M$. Then, use a \textit{softmax} function to transform logit scores into the categorical distribution, as
\begin{align}
\mathrm{Pr}(\hat m_i = m|\mathcal{H}(t)) = \mathrm{softmax}(\mathrm{logit}(\hat m_i))_m
\end{align}
where $\mathrm{softmax}(\mathrm{logit}(\hat m_i))_m$ means choose the $m$-th element after \textit{softmax}'s output.

In this way, a cross-entropy loss for categorical classification will be added to the log-likelihood term given the true event type of the $i$-th event, to maximize the joint likelihood of both next event arrival time and type which is regarded as independent.
There are also works on maximizing joint likelihood in conditional forms, i.e. time conditioned on marks \cite{enguehard2020neural, thp} and marks conditioned on time \cite{shchur2021neural}, which can further capture the dependencies between time and event type. In our further study, we unify all the methods as independent modeling for time and event. 
\section{Deep Temporal Point Process Extension}
The necessary parts of deep temporal point process are history encoder and intensity function. And here we conclude these two parts according to a majority of existing methods. Besides, we modify and extend them further.
\subsection{Extensive Historical Event Encoders} 
\subsubsection{Recurrent-based encoders}\label{sec:2.2}
RNN units including GRU and LSTM have all been used as a history encoder \citep{rmtpp, fnn, lognorm}, i.e.
\begin{align}
\bm{h}_0 = \bm{0};\quad \bm{h}_{i} = \mathrm{RNN}(\bm{e}_{i-1}, \bm{h}_{i-1})
\end{align}
The advantage of this history encoder is that it occupies low storage space, due to serial computing. However, the serial computing limits the computational speed, in both 'forward' and 'backward' processes, and probably compromises the performance, which is resulted from gradient vanishing effects and long-term memory loss \cite{gradientvanish} theoretically.
\subsubsection{Attention-based encoders}\label{sec:3.1.2}
For recurrent encoders' slow serial computing and loss of long-term information, self-attention \citep{sahp} is proposed with fast parallel computing and capability of encoding long-term sequences, which can be written as:
\begin{align}
    \bm{h}_i 
    &= \sum_{j=1}^{i-1} \phi(\bm{e}_j, \bm{e}_{i-1}) \psi(\bm{e}_j) / \sum_{j=1}^{i-1} \phi(\bm{e}_j, \bm{e}_{i-1}), \label{eq:attmecha}
\end{align}
where $\phi(\cdot,\cdot)$ maps two embedding into a scalar called attention weight, and $\psi$ transforms $\bm{e}_{j}$ into a series of $D$-dimensional vectors called values. While the attention mechanism solves some of the problems existing in recurrent-based encoders, the $O(N^2)$ space complexity caused by the attention matrix also brings about trouble when the sequence is very long.

\subsubsection{Fourier transform encoders} In addition, we \textbf{adapt recently proposed Fast Fourier Transform} (FFT) module in natural language processing \citep{Fnet} to our history encoder family, which aims speed up the computation and replace attention mechanism. 
\begin{align}
    \bm{h}_i 
    &= \text{Top}_k\left\{\mathrm{FFT}([\mathrm{FFT}(\bm{e}_1); \mathrm{FFT}(\bm{e}_2); \ldots;\mathrm{FFT}(\bm{e}_{i-1})])\right\}.
\end{align}
The $\mathrm{FFT}(\cdot)$ represents the Fast Fourier Transformation, which firstly operates on the events' embedding, then on the whole sequence, and $\text{Top}_k\{\cdot\}$ means choosing the highest $k$ frequencies in the set as the history encoding. Note that the dimension of $\bm{e}_j$ has to equal to $D$ in this way. The Fourier transform encoders can also capture long-term patterns due to the global property of the sequences' spectrum. Here we explain why we choose the top $k$ frequencies. The lengths of history sequence of event embedding are not equal, so the padding operation is necessary for batch processing. In this way, lots of sequences contain a large number of the same padding values, which are useless information and lead to low-frequency values in spectral. Therefore, it is reasonable to filter the low frequency and capture the high frequency, because the latter one contains more information of the historical sequence.

The Fourier tansform encoders enjoy fast computational time complexity of $O(N\log N)$, and can also capture the global feature of sequences. However, the 'backward' process of gradient propagation leads to huge memory costs in implementation.
\subsection{Extensive Conditional Intensity Functions}
We modify a line of conditional intensity functions and extend the mixture distribution family with several distributions supported on the half-interval. Note that all the considered CIF has the closed-form integral for maximum likelihood estimation. For those who are not computational manageable, we try to modify it with only tiny changes to make the integral closed, and for those not included below, we do not think of them as revisable.   For simplicity of notation, we omit the subscript $m$ when describing the single event's distribution. 
\subsubsection{Modified Fully Neural Network Intensity}
\label{sec:3.2.1}
Omi et al. \citep{fnn} directedly fit the integrated CIF which is called cumulative harzard function (CHF) as $\Lambda^*(t) = \int^t_{t_{i-1}}\lambda^*(s)ds$ for $t\in[t_{i-1}, t_i)$ by a 3-layers fully connected feedforward neural network 
\begin{align}
\Lambda^*(t) = \text{softplus}(\bm{W}^{(3)}\text{tanh}&(\bm{W}^{(2)}\text{tanh}(\bm{W}_t^{(1)}[t;\bm{h}_i] + \bm{b}^{(1)})\notag\\ & + \bm{b}^{(2)}) + \bm{b}^{(3)}), \label{eq:fnnint}
\end{align}
where all the elements in $\{\bm{W}^{(m)}\}_{1\leq m \leq 3}$ are non-negative weights and $\{\bm{b}^{(m)}\}_{1\leq m \leq 3}$ are bias terms. The CIF can be obtained by $\lambda^*(t) = \frac{\partial}{\partial t}\Lambda^*(t)$. However, Oleksandr Shchur et al. \citep{lognorm} pointed out that the saturation of $\text{tanh}(\cdot)$ leads that $\lim_{t \rightarrow \infty}\Lambda^*(t) < \infty$. As a result, the PDF of the distribution does not integrate to 1. We modify it by adding a positive term, and thus the integrated CIF reads
\begin{align}
        \Lambda^*(t) = \text{softplus}(\bm{W}^{(3)}\text{tanh}&(\bm{W}^{(2)}\text{tanh}(\bm{W}_t^{(1)}[t;\bm{h}_i] + \bm{b}^{(1)})\notag \\
        &+ \bm{b}^{(2)}) + \bm{b}^{(3)}) + b_tt, \label{eq:fnnintmodify}
\end{align}
where $b_t > 0$. Thus the problem of 'non-infinity' is solved. Because it use fully neural network to model the integral of CIF, we call it \textit{FNNIntegral}. 

There are also following works on universal approximation to the CIF. For example, Oleksandr Shchur et al. \cite{triangular} further used very flexible triangular maps to approximate a line of temporal point processes, and Alexander Soen \cite{unipoint} give approximation analysis theoretically, and conduct empirical study on a series of basis functions as universal approximators for the target intensity functions.
\subsubsection{Mixture Family Distribution}
\label{sec:3.2.2}
We briefly introduce mixture distribution family to approximate the target PDF governed by CIF. According to Theorem 33.2 in \cite{Asymptotictheory}, the translated and dilated mixture distribution can be universal approximation for any continuous density on $\mathbb{R}$, so we extend the Log-normal Mixture to an extensive mixture family distribution. 
\\
\\
\textbf{Log-normal Mixture.} Oleksandr Shchur et al. \citep{lognorm} proposed to use the mixture of Log-normal to approximate any distribution. The feasible computation of its PDF and cumulative distribution function (CDF) decides the closed-form of CIF and CHF as its integral, by
\begin{align}
    \lambda_{LN}^*(t) = \frac{f_{LN}^*(t)}{1-F_{LN}^*(t)}; \quad 
    \Lambda_{LN}^*(t) = -\ln ({1-F_{LN}^*(t)}), \label{eq:cifandcdf}
\end{align}
where $F_{LN}^*(t)$ is CDF and $f_{LN}^*(t)$ is the PDF. And the mixture form reads 
\begin{align}
 f_{LNM}^*(t) = \sum_{k=1}^K w_k\frac{1}{t\sigma_k\sqrt{2\pi}} &\exp(-\frac{(\ln (t-t_{i-1}) - \mu_k)^2}{2\sigma_k^2}),\label{eq:cdflognorm}
\end{align}
for $t\in[t_{i-1}, t_{i})$, $K$ are the mixture distribution numbers, $\{w_k\}_{1\leq k \leq K}$ are the non-negative mixture weights, with $\sum_{k=1}^Kw_k = 1$, $\sigma_k > 0$ for any $k$, and 
\begin{align}
    \bm{\Theta}(t) = \{w_k(t), \mu_k(t), \sigma_k(t)\}_{1\leq k \leq K} = \bm{\chi}(\bm{h}_i) 
\end{align}
for $t\in[t_{i-1}, t_{i})$, where $\bm{\chi}(\bm{h}_i) = \bm{W}_\theta \bm{h}_i + \bm{b}_\theta$ and $\bm{W}_\theta \in \mathbb{R}^{3K \times D}$, $\bm{b}_\theta \in \mathbb{R}^{3K}$ .
Although the CDF has no closed form, the approximation of \textit{erf} function has minor deviation and permits gradient back-propagation, allowing both 'forward' and 'backward' process.  More details on implementation see Appendix B.1.
\\ 
\\
\textbf{Gompertz Mixture.} 
Du et al. \citep{rmtpp} modeled the CIF with the corresponding the PDF as 
\begin{align*}
    \lambda_{GP} ^*(t) &= \exp(\beta (t-t_{i-1}) + \bm{v}^T\bm{h}_i + b_t),\\
\end{align*}
which is a Gompertz distribution, whose PDF reads 
\begin{align}
    f_{GP}^*(t) &= \eta  \exp(\beta  (t-t_{i-1}) - \frac{\eta}{\beta} (\exp(\beta  (t-t_{i-1})) - 1)),
\end{align}
for $t\in[t_{i-1,t_i})$, where $\eta = \exp(\bm{v}^T\bm{h}_i + b_t)$, and $\beta > 0$. Note that the process degenerates to Possion when $\beta = 0$ according to the CIF. We extend the distribution as a mixture as the last paragraph just introduced, which is
\begin{align}
    f_{GPM}^*(t) = \sum_{k=1}^K w_k\eta_k  &\exp(\beta_k (t-t_{i-1})\notag\\ 
                                    &- \frac{\eta_k}{\beta_k} (\exp(\beta_k  (t-t_{i-1})) - 1)), \label{eq:gomppdf}
\end{align}
 for $t\in[t_{i-1}, t_{i})$, where $\beta_k > 0$ and $\eta_k> 0$ for any $k$. The parameters are obtained as a function of history encoding $\bm{h}_i$, that is for $t\in[t_{i-1}, t_{i})$, $\bm{\Theta}(t) = \{w_k(t), \beta_k(t), \eta_k(t)\}_{1\leq k \leq K} = \bm{\chi}(\bm{h}_i)$. More details on implementation see Appendix B.2.
\\
\\
 \textbf{Exp-decay Mixture.} Zhang et. al. \citep{sahp} modeled the intensity function as exponential-decaying form which is similar to classical Exp-decay Hawkes Process, except that a nonlinear transform $\textit{softplus}$ is stacked after, which causes the integral term computationally unmanageable. We remove the final transformation of non-linearity, and obtain the CIF of defined Exp-decay distribution as
 \begin{align}
    \lambda_{ED}^*(t) = \eta  \exp(-\beta (t-t_{i-1})) + \alpha
 \end{align}
 where $\alpha$ is the basic intensity and the first term indicates the impacts of historical events decay with an exponential ratio.
 By using the distribution as component, we propose the mixture of Exp-decay distribution, whose PDF reads
 \begin{align}
    f_{EDM}^*(t) = \sum_{k=1}^Kw_k(&\eta_k  \exp(-\beta_k (t-t_{i-1})) + \alpha_k)\notag\\ 
    &\exp((\frac{\eta_k}{\beta_k} - 1) \exp(-\beta_k(t-t_{i-1})) \notag\\
    &- \alpha_k (t-t_{i-1})),
\end{align}
 for $t\in[t_{i-1}, t_{i})$, whose parameters are all positive, calculated by $ \bm{\chi}(\bm{h}_i) = \{w_k(t), \alpha_k(t), \beta_k(t), \eta_k(t)\}_{1\leq k \leq K}$.  Appendix B.3 gives further implementation details. 
\\
\\
\textbf{Other mixture distributions.} Besides, we extend other mixture of distributions as choices, the components of which include Weibull \citep{weibull} whose CIF reads $\lambda^*_{WB}(t) = \eta  \beta (\eta  (t-t_{i-1}))^ {\beta - 1}$, and Log-Cauchy which is used to model a 'super-heavy tailed' distribution \citep{logcauchy} with the PDF written as $f_{LC}^*(t) = \frac{1}{(t-t_i) \pi}  \frac{\sigma}{(\ln(t-t_i) - \mu)^2 + \sigma ^ 2}$. Gaussian mixture distribution is also included, although the support is not the positive half real line. More details are showed in Appendix B.4 and B.5 respectively.

\subsubsection{Normalizing Flows}
Oleksandr Shchur et al. also proposed methods on modeling temporal point process based on normalizing flows \cite{lognorm}. The PDF of the target process is approximated by the transofrmation of a known multivariate distribution, e.g. Gaussian, according to formula of change of variables, which reads
\begin{align}
    f(t) = |\mathrm{det}(J_{G})|\tilde{f}(G(t)) = |\frac{\partial}{\partial t}G(t)|\tilde{f}(G(t))
\end{align}
where the PDF of $t$ is $f(\cdot)$, the PDF of $z = G(t)$ is $\tilde{f}(\cdot)$ which is predefined, and $G(\cdot)$ is a invertible mapping which the model tries to learn.
During training, $t = (t_1, \ldots, t_N)$ is fed into the model to maximize the likelihood, and in sample generating process, $t$ is generated by first sampling $z$ with the known distribution, and the target sequence is obtained with $G^{-1}(z)$.
\\
\\
\textbf{DSFlow and SOSFlow.} The deep sigmoidal flow (DSF) from \cite{huangnaf} and Sum-of-squares (SOS) polynomial flow from \cite{jaini2019sumofsquares} can be employed to approximate the $f(t)$, where the single layer is 
\begin{align}
    G_{DSF}^{-1}(z) &= \mathrm{sigmoid} (\sum_{k=1}^K\omega_k\cdot\mathrm{sigmoid}(\frac{z - \mu_k}{\sigma_k}));\\
    G_{SOS}^{-1}(z) &= \omega_0 + \sum_{k=1}^K\sum_{p=1}^{R}\sum_{q=1}^R \frac{\omega_{p,k} \omega_{q,l}}{p+q+1} z^{p+q+1}.
\end{align}
In the training process, the time interval $\tau$ is first transformed by $\log(\cdot)$ to convert a positive value into $\log(\tau) \in \mathbb{R}$. Then, multiple layers as variable transformation are stacked, in the final of which a $sigmoid$ is used to map the value into $[0,1]$.
All the parameters are obtained by a function of history encoding like the previously discussed. 

\section{Case study}
\label{sec:4}
In this part, we give an introduction to recently proposed representative models of deep temporal point process. We dismantle them into the four parts as discussed. The history encoder and conditional intensity function are necessary for modeling the process. The relational discovery parts have not drawn lots of research interests until nearly one year ago. For learning approaches, the first eight methods all use maximum likelihood estimation with stochastic gradient descent as their learning approaches.
Further, several learning approaches other than MLE with SGD are introduced, including reinforcement learning, adversarial\&discriminativelearning and noise contrastive learning.
A snapshot of the first eight representative methods is provided in Table. \ref{tab:review}.
\subsection{Recurrent Marked Temporal Point Process}
Recurrent Marked Temporal Point Process (RMTPP) \cite{rmtpp}, which is, to our knowledge, the first deep-learning-based method to model temporal point process, achieves tremendous improvements in both fitting and prediction performance compared with the classical point process models.   

It firstly embeds time through linear transformation, and event type through token embedding technique as Eq. \ref{eq:eventemb}, and then the recurrent neural network (RNN) \citep{rnn} serves as the history encoder to embed history into vectors. 
The CIF in RMTPP is the same as it in Gompertz distribution \citep{gompertz} with a closed-form of integral, which is a special case of the Gompertz Mixture distribution (as discussed in Sec. \ref{sec:3.2.2}) when mixture component number equals one.
The optimization target is joint likelihood of both event time and type, and SGD based optimizer is used.
\subsection{Event Recurrent Temporal Point Process}
Event Recurrent Point Process (ERTPP) \cite{erpp} further modifies RMTPP. 
On one hand, it explores different neural network structure's effects on model's performance, by using LSTM \cite{lstm} to model the series, and fuses different information through multi-LSTM structure to enhance history encoder's expressivity.
On the other hand, the CIF it used is a single Gaussian distribution. In this way, when the variance of Gaussian distribution is predefined, maximizing the log-likelihood is equivalent to minimizing the mean square error (MSE), i.e.
\begin{align}
    l(\bm{\Theta}) = \frac{1}{\sigma^2}\sum_{i=1}^{N}(t_i - \mu(t)_{m_i})^2,
\end{align}
where $\bm{\Theta}(t) = \mu(t)$ and $\mu(t)_{m_i} = \bm{\chi}(\bm{h}_i)_{m_i}$, the subscript in which means selecting the $m_i$-th component of the parameter $\mu$, representing the $m_i$ type's mean value; $\sigma$ is not learnable, and is set as $\sigma^2 = 10$ in the implementation.

An obvious problem of the used CIF is that the support set is not half positive real line, so in the process of generating samples, invalid negative time intervals may be generated by the model with a small probability.  
\subsection{Continuous Time Neural Point Process}
Continuous Time Neural Point Process, which is also called Continuous Time LSTM (CTLSTM) \cite{ctlstm}, proposes a history encoder by using LSTM  as recurrent units and introduces a temporal continuous memory cell in it.
The history encoder has some differences from the concluded forms in Sec.~\ref{sec:2.2}, as it allows a continuous time interval $\tau$ as recurrent encoders' input, as follows
\begin{align}
    \bm{h}_0 = \bm{0};\quad \bm{h}_{i} = \mathrm{CTLSTM}(\bm{e}_{i-1}, \bm{h}_{i-1}, \tau_{i-1}),
\end{align}  
where $\bm{e}_{i-1}$ only contains information of event types.
The CIF is modeled as the summation of decaying exponential functions, thus considering all the historical impacts, while a \textit{softplus} function is used for non-linearity transformation.
Because of the non-linear function stacked in the end, the CIF in CTLSTM has no closed-form integral, and thus this term is approximated by Monte Carlo stochastic integration methods, which dramatically increase the computational cost. 

In recent years, lots of continuous-time-dependent models are proposed, and also share some common ideas with CTLSTM, such as exponential-time-decayed RNN \cite{cao2018brits}, neural-ODE-based time series model \cite{rubanova2019latent} and neural spatio-temporal point process \cite{chen2021neural}.
\subsection{Fully Neural Network Point Process}
The history encoder used in Fully Neural Network Point Process (FNNPP) \cite{fnn} is also recurrent-based. The novelty of the work is that it proposes a fully connected neural network as a general approximator to directly model the CHF which is the integral term of the CIF as its output, and the derivative of the output with respect to time interval is regarded as the CIF. 
In this formulation of CIF, the difficult problem raised by integration is turned into a differentiation problem, which is much easier to handle. The model adopts an expressive non-parameterized neural network for statistical inference and assures computational manageability as well.
In Sec. \ref{sec:3.2.1}, we give a brief introduction to the CIF formulation. A more detailed form is that, for $t\in[t_{i-1},t_i)$
\begin{align}
    \Lambda^*_m(t) &= \int_{t_{i-1}}^{t}\lambda^*_m(s)ds = \mathrm{MLP}(t);\\
    \lambda^*_m(t) &= \frac{\partial \Lambda^*_m(t)}{\partial t},
\end{align}
where MLP($\cdot$) is a multi-layer feedforward neural network, originally modeled in Eq. \ref{eq:fnnint} form, and the log-likelihood reads
\begin{align}
    l(\bm{\Theta}) &= \sum_{i=1}^N \left [\log \frac{\partial \Lambda^*_{m_i}(t)}{\partial t}\middle |_{t=t_i} - \Lambda^*_{m_i}(t_i)\right ].
\end{align}
The proposed CIF has both powerful expressivity as well as closed integral, showing good performance in fitting a variety of synthetic temporal point processes, including renewal, self-correcting, Hawkes process and so on.
However, there exists problem of 'non-infinity' of the CIF in Eq. \ref{eq:fnnint}, so we use a trivial trick to modify it as shown in Eq. \ref{eq:fnnintmodify}. 
\subsection{Log-normal Mix Point Process}
While the history encoder is still recurrent-based, LogNormMix \cite{lognorm} proposes to use a mixture of Log-normal distribution (as discussed in Sec. \ref{sec:3.2.2}) to approximate the PDF of the process, according to the following universal approximation theorem of mixture model.
\\
\textbf{\textit{Theorem 1.}} (Theorem 33.2 in \cite{Asymptotictheory})
Let $p(x)$ be a continuous density on $\mathbb{R}$. If $q(x)$ is any density on $\mathbb{R}$ and is also continous, then given $\epsilon >0$, and a compact set $\mathcal{S} \in \mathbb{R}$, there exist number
of components $K \in \mathbb{N}$,  mixture coefficients $\bm{w} \in \Delta^{K-1}$, locations $\bm{\mu} \in \mathbb{R}^{K}$, and scales $\bm{s} \in \mathbb{R}_+^{K}$, s.t. for the mixture distrubution $ \hat p(x) = \sum _{k=1}^K w_l \frac{1}{s_l} q( \frac{x - \mu_l }{s_l})$, it holds $\sup_{x\in\mathcal{S}}|p(x) - \hat p(x)| < \epsilon$.

The core idea of the work is that modeling $f^*(t)$ instead of $\lambda^*(t)$ does not impose any limitation when the closed-form of $f^*(t)$ is given. Moreover, as the mixture distribution can also achieve great expressivity, using the mixture of PDF with positive support is flexible. 
In this way, the proposed formulation of CIF possesses the two good properties: computational manageability and powerful expressivity.

Besides, the mixture of closed-form distribution permits simpler data generation and further statistical analysis.
The convenient sample generation of LogNormMix enables to learn the true underlying data distribution even if the data is missing, through the data imputation based on sampling the missing values in training process.
\subsection{Self-attentive Hawkes Process}
Self-attentive Hawkes Process (SAHP) \cite{sahp} proposes a multi-head attention network as the history encoder (as discussed in Sec. \ref{sec:3.1.2}), to capture the long-term patterns as well as speed up the computation.
As the pure attention may compromise the performance \cite{dong2021attention}, the feedforward layer containing batch-normalization and residual connection is stacked after, leading the network units to be very similar to Transformer \cite{attention}.
In addition, the expected attention weights between two types of events can be calculated according to attention matrix, which is considered as a relational interpreter in SAHP.

Besides, another contribution of it is that the time embedding called time shifted positional encoding in Eq. \ref{eq:eventemb} is trigonometric-based, utilizing the positional encoding methods in natural language process \cite{zheng2021rethinking}, as
\begin{align}
    \bm{\omega}(t_i) = [\sin(\omega_1 i + \omega_2 t_i); \cos(\omega_1 i + \omega_2 t_i)],
\end{align} 
where $\omega_1 i$ is positional term, and $\omega_2 t_i$is shift term. $\omega_2$ is learnable parameters, and $w_1$ is predefined.

The formulation of CIF reads
\begin{align}
    \lambda_m^*(t) = \mathrm{softplus}(\eta \exp(-\beta (t-t_{i-1})) + \alpha),
\end{align}
because the \textit{softplus} is employed to constrain the CIF to be positive, the parameters in the CIF can be both positive and negative. We classify it as a special single Exp-decay distribution (as discussed in Sec. \ref{sec:3.2.2}), except that a nonlinear function follows in the final layer. The $softplus$ causes that the model requires the Monte Carlo integration methods to estimate the CHF in the log-likelihood term, and increase the computational cost dramatically.

\subsection{Transformer Hawkes Process}
Transformer Hawkes Process (THP) \cite{thp} also proposes to use multi-head attention mechanism with Transformer units to construct the history encoder. Trigonometric time encoding is also used, while there is no positional term and learnable parameters. 

CIF used in the model is different from the former, employing a continuous time-dependent neural network, which reads
\begin{align}
    \lambda_m^*(t) = \mathrm{softplus}(\alpha_m\frac{t-t_{i-1}}{t_{i-1}} + \bm{\chi}(\bm{h}_{i-1})),
\end{align}
where $\bm{\chi}(\cdot)$ is a linear transformation. When the \textit{softplus} is removed, the CIF will not keep positive, and the first-order polynomial with respect to time cannot be extended as an intensity function. Therefore, we do not include it for further comparison.
The integral term still has no closed-form, so numerical or Monte Carlo integration is used.

A latent graph structure is established according to the similarity of pairwise event embeddings. The obtained similarity matrix describing the adjacency of relational graph of event types is first used in the attention mechanism as
\begin{align}
    \phi_{m,m'}(\bm{e}_j, \bm{e}_{i-1}) = \bm{A}_{m,m'} + \phi(\bm{e}_j, \bm{e}_{i-1}), \label{eq:graphatten}
\end{align}
where $\bm{A}_{m,m'} = (\bm{E}^T\bm{m})^T \bm{\Omega} (\bm{E}^T\bm{m}') \in \mathbb{R}$, $(\bm{E}^T\bm{m})$ is the event embedding as discussed in Eq. \ref{eq:eventemb} and $\bm{\Omega}$ is a learnable metric matrice. 
And when the prior graph edge set $\bm{\mathcal{E}}_{\mathrm{pri}}$ is given, it also serves as a regularization term that encourages the similarity to be large when there exists an edge between $m$ and $m'$, which reads
\begin{align}
    l_{\mathrm{graph}} = \sum_{m=1}^{M} &\sum_{m'=1}^{m} -\log(1-\exp((\bm{E}^T\bm{m})^T \bm{\Omega} (\bm{E}^T\bm{m}')))\notag\\
                                                        &+\mathbbm{1}_{(\bm{m},\bm{m}')\in \bm{\mathcal{E}}_{\mathrm{pri}}}((\bm{E}^T\bm{m})^T \bm{\Omega} (\bm{E}^T\bm{m}')).
\end{align}
This regularization means if two vertices are connected in graph, then the
regularizer will promote attention between them, and vice versa.
Besides, in the setting of next event prediction, it utilizes 'time conditioned on event types', enabling to further model the dependencies between time and event types.
\subsection{Dependent Graph Neural Point Process}
Dependent Graph Neural Point Process (DGNPP)\cite{dgnpp} follows most of settings in SAHP including history encoders and formulation of CIF, while what it aims to explore is the complex relation among different types through a graph structure learning method. 
The graph is generated based on a random graph theory, i.e. Erdös–Rényi model. And the generated graph adjacency matrix $\bm{A}_{m,m'}$ which measures impacts from type $m$ to type $m'$, will be used as a mask on the attention matrix from one type to another, which refactorized the Eq. \ref{eq:attmecha} by
\begin{align}
    \phi_{m,m'}(\bm{e}_j, \bm{e}_{i-1}) = \bm{A}_{m,m'}\phi(\bm{e}_j, \bm{e}_{i-1}).
\end{align}
What differs from Eq. \ref{eq:graphatten} is that the $\bm{A}_{m,m'} \in \{0,1\}$, which is generated by a discrete distribution.
Graph generator is based on Bernoulli distribution, which is parametrized by the inner product of embeddings of event type. Gumbel-Max \cite{gumbel1, gumbel2} trick is used to both draw samples and propagate gradient. 

Besides, it proposes a bilevel programing as learning approaches, in which the validation set is used to tune the parameters to find the optimal graph, and training set is to maximize the log-likelihood of the fixed graph structure.
To develop an efficient algorithm to find the approximated solutions of bilevel learning, it proposes a learning dynamics based on iterative gradient descents, which guarantees the convergence of the bilevel programming \cite{shaban2019truncated}.

\begin{table*}[]
    \resizebox{1\linewidth}{!}{
        \begin{tabular}{l|ll|ll|l}
        \toprule
        Methods                      & \multicolumn{2}{c}{Workflows}                     & \multicolumn{2}{c}{Properties}                          & \multicolumn{1}{c}{Released Codes}                                                                                                                   \\
        \midrule
        \multirow{4}{*}{RMTPP\cite{rmtpp}}      & History Encoder:      & RNN                        & Marked Modeling:         & \CheckmarkBold & \multirow{4}{*}{\begin{tabular}[c]{@{}l@{}}{[}Tensorflow{]}:\\ \url{https://github.com/musically-ut/tf\_rmtpp}\end{tabular}}              \\ \cline{2-5}
                                    & Intensity Function:   & Gompertz                   & Closed-from Likelihood:  & \CheckmarkBold &                                                                                                                                     \\ \cline{2-5}
                                    & Relational Discovery: & /                          & Closed-from Expectation: & \XSolidBrush  &                                                                                                                                     \\ \cline{2-5}
                                    & Learning Approaches:  & MLE with SGD               & Closed-from Sampling:    & \CheckmarkBold &                                                                                                                                     \\
        \midrule
        \multirow{4}{*}{ERTPP\cite{erpp}}      & History Encoder:      & LSTM                       & Marked Modeling:         & \CheckmarkBold & \multirow{4}{*}{\begin{tabular}[c]{@{}l@{}}{[}Tensorflow{]}:\\ \url{https://github.com/xiaoshuai09/Recurrent-Point-Process}\end{tabular}} \\ \cline{2-5}
                                    & Intensity Function:   & Gaussian                   & Closed-from Likelihood:  & \CheckmarkBold &                                                                                                                                     \\ \cline{2-5}
                                    & Relational Discovery: & /                          & Closed-from Expectation: & \CheckmarkBold &                                                                                                                                     \\ \cline{2-5}
                                    & Learning Approaches:  & MLE with SGD               & Closed-from Sampling:    & \XSolidBrush   &                                                                                                                                     \\
        \midrule
        \multirow{4}{*}{CTLSTM\cite{ctlstm}}     & History Encoder:      & (CT)LSTM                   & Marked Modeling:         & \CheckmarkBold & \multirow{4}{*}{\begin{tabular}[c]{@{}l@{}}{[}Theano{]}:\\ \url{https://github.com/HMEIatJHU/neurawkes}\end{tabular}}                     \\ \cline{2-5}
                                    & Intensity Function:   & Exp-decay + softplus      & Closed-from Likelihood:  & \XSolidBrush   &                                                                                                                                     \\ \cline{2-5}
                                    & Relational Discovery: & /                          & Closed-from Expectation: & \XSolidBrush   &                                                                                                                                     \\ \cline{2-5}
                                    & Learning Approaches:  & MLE with SGD               & Closed-from Sampling:    & \XSolidBrush   &                                                                                                                                     \\
        \midrule
        \multirow{4}{*}{FNNPP\cite{fnn}}      & History Encoder:      & LSTM                       & Marked Modeling:         & \XSolidBrush   & \multirow{4}{*}{\begin{tabular}[c]{@{}l@{}}{[}Keras{]}:\\ \url{https://github.com/omitakahiro/NeuralNetworkPointProcess}\end{tabular}}                      \\ \cline{2-5}
                                    & Intensity Function:   & FNNIntegral (as CHF)               & Closed-from Likelihood:  & \CheckmarkBold &                                                                                                                                     \\ \cline{2-5}
                                    & Relational Discovery: & /                          & Closed-from Expectation: & \XSolidBrush   &                                                                                                                                     \\ \cline{2-5}
                                    & Learning Approaches:  & MLE with SGD               & Closed-from Sampling:    & \XSolidBrush   &                                                                                                                                     \\
        \midrule       
        \multirow{4}{*}{LogNormMix\cite{lognorm}} & History Encoder:      & LSTM                       & Marked Modeling:         & \CheckmarkBold & \multirow{4}{*}{\begin{tabular}[c]{@{}l@{}}{[}Pytorch{]}:\\ \url{https://github.com/shchur/ifl-tpp}\end{tabular}}                         \\ \cline{2-5}
                                    & Intensity Function:   & Log-norm Mixture           & Closed-from Likelihood:  & \CheckmarkBold &                                                                                                                                     \\ \cline{2-5}
                                    & Relational Discovery: & /                          & Closed-from Expectation: & \CheckmarkBold &                                                                                                                                     \\ \cline{2-5}
                                    & Learning Approaches:  & MLE with SGD               & Closed-from Sampling:    & \CheckmarkBold &                                                                                                                                     \\
        \midrule
        \multirow{4}{*}{SAHP\cite{sahp}}       & History Encoder:      & Transformer                & Marked Modeling:         & \CheckmarkBold & \multirow{4}{*}{\begin{tabular}[c]{@{}l@{}}{[}Pytorch{]}:\\ \url{https://github.com/QiangAIResearcher/sahp\_repo}\end{tabular}}           \\ \cline{2-5}
                                    & Intensity Function:   & Exp-decay + softplus          & Closed-from Likelihood:  & \XSolidBrush   &                                                                                                                                     \\ \cline{2-5}
                                    & Relational Discovery: & Attention Matrix           & Closed-from Expectation: & \XSolidBrush   &                                                                                                                                     \\ \cline{2-5}
                                    & Learning Approaches:  & MLE with SGD               & Closed-from Sampling:    & \XSolidBrush   &                                                                                                                                     \\
        \midrule
        \multirow{4}{*}{THP\cite{thp}}        & History Encoder:      & Transformer                & Marked Modeling:         & \CheckmarkBold & \multirow{4}{*}{\begin{tabular}[c]{@{}l@{}}{[}Pytorch{]}:\\ \url{https://github.com/SimiaoZuo/Transformer-Hawkes-Process}\end{tabular}}   \\ \cline{2-5}
                                    & Intensity Function:   & Linear function + softplus       & Closed-from Likelihood:  & \XSolidBrush   &                                                                                                                                     \\ \cline{2-5}
                                    & Relational Discovery: & Structure learning         & Closed-from Expectation: & \XSolidBrush   &                                                                                                                                     \\ \cline{2-5}
                                    & Learning Approaches:  & MLE with SGD               & Closed-from Sampling:    & \XSolidBrush   &                                                                                                                                     \\
        \midrule
        \multirow{4}{*}{DGNPP\cite{dgnpp}}      & History Encoder:      & Transformer                & Marked Modeling:         & \CheckmarkBold & \multirow{4}{*}{No released code until now.}                                                                                        \\ \cline{2-5}
                                    & Intensity Function:   & Exp-decay + softplus         & Closed-from Likelihood:  & \XSolidBrush   &                                                                                                                                     \\ \cline{2-5}
                                    & Relational Discovery: & Bilevel structure learning & Closed-from Expectation: & \XSolidBrush   &                                                                                                                                     \\ \cline{2-5}
                                    & Learning Approaches:  & MLE with SGD               & Closed-from Sampling:    & \XSolidBrush   &                                                                                                                                    
        \\\bottomrule
        \end{tabular}
    }
    \\
    \caption{A conclusion of the eight representative methods. 
    In \textit{Properties}, \textit{Marked Modeling} means if the original model can handle the task of next event prediction, \textit{Closed-form Likelihood, Expectation and Sampling} means if the distribution governed by the conditional intensity has closed-from likelihood for optimization, closed-from expectation on time for next arrival time prediction and closed-from sampling for sequence generation.}\label{tab:review}
\end{table*}
\subsection{Reinforcement learning}
\label{sec:4.9}
Now we start to give introduction on several works focusing on learning approaches which were different from maximum log-likelihood with SGD.
\\
\\
The application of reinforcement learning to temporal point process \cite{reinforcement1} is based on two aspects:
\begin{itemize}
    \item The agent’s actions and environment’s feedback are asynchronous stochastic events in continuous time.
    \item The policy is a conditional intensity function and a mark categorical distribution, which is used to
    sample the times and marks of the agent’s actions.
\end{itemize}

The angent is defined as $p^*_{\mathcal{A},\theta_1} = \lambda^*_m(t|\bm{\Theta}_1)$, and the environment’s feedback is $p^*_{\mathcal{F},\theta_2} = \lambda^*_m(t|\bm{\Theta}_2)$.
Therefore, on a stochatic reward $R(\cdot)$, the target to maximize is 
\begin{align}
    \max _{p^*_{\mathcal{A},\theta_1}} \mathbb{E}_{\mathcal{A}_t \sim p^*_{\mathcal{A},\theta_1}, \mathcal{F}_t \sim p^*_{\mathcal{F},\theta_2}}[R(t)].
\end{align}

The policy which is the marked CIF is formulated as RMTPP does, while the sampling action of the Gompertz distribution is developed to get the next action time.

Another reinforcement-based learning approach \cite{reinforcement2} utilizes some perspective from adversarial learning, trying to generate samples from the generative model and monitor the quality of the samples in the process of training until the samples and the real data are indistinguishable. It uses a more flexible RNN to gradually improve the policy, and inverse reinforcement learning formulation to uncover the reward function. 
\subsection{Adversarial and discriminativelearning}
It is claimed that using the adversarial and discriminativelearning methods \cite{adversarial} can further improve the maximum likelihood estimation.
The training strategy is also a bilevel dynamics, including discriminativeprocess and adversarial process.
\begin{itemize}
    \item Discimitive process: To turn the task into a classification one, the method first split limited time interval for several small intervals. By using a discriminativeloss mesauring the difference of event counts between ground truth and the generated timestamps in each interval, it optimize the generator's parameters, i.e. the model to fit the process.   
    \item Adversarial process: An adversarial critic is established to measure the Wasserstein distance \cite{arjovsky2017wasserstein} between distribution of generated sequence and ground truth. The generator's parameter is further update to decrease the defined distance, while the critic tries to distinguish the generated from the ground truth. 
\end{itemize}
The motivation of this work is that most model obtained by MLE-based learning usually fits the sequence well, but prediction capability is limited. Therefore, the discriminativeprocess is used to improve the predictive performance.               
\subsection{Noise contrastive learning}
Because the maximum likelihood estimation may suffer from intractable integral term, a learning strategy proposed in \cite{noisecontrast} employes noise contrastive estimation \cite{guntmann2010} for learning a temporal point process model.
Parameters of the model are learned by solving a binary classification problem where samples are classified into two classes, namely true sample or noise sample. And the target is to maximize
\begin{align}
    \mathbb{E}_{t \sim f^*_d}[\log \mathrm{Pr}(y=1|t)] + N_n\mathbb{E}_{t \sim f^*_n}[\log\mathrm{Pr}(y=0|t)],
\end{align}
where $ \mathrm{Pr}(y=1|t)$ denotes the probability that the event time $t$ is a sample observed in data distribution $f*_d$,
$ \mathrm{Pr}(y=0|t)$ denotes the probability that the event time $t$ is not observed in the data but generated from the noise distribution $f^*_n$,
and $N_n$ is the number of noise samples generated for each sample in the data.
The distribution of the process as $f^*_d$ is approximated by the neural networks, and the $\mathrm{Pr}(y|t)$ is formulated as 
\begin{align}
    \mathrm{Pr}(y|t) = \left[\frac{ f^*_d(t)}{f^*_d(t) + N_nf^*_n(t)}\right]^{y}\left[\frac{f^*_n(t)}{f^*_d(t) + N_nf^*_n(t)}\right]^{1-y},
\end{align}
which is a Bernoulli distribution.

A recently proposed noise contrastive framework \cite{noisecontrast2} further adapts the learning strategy to the continuous-time scenarios. As a more general idea, it has provable guarantees for optimality, consistency and efficiency, making contributions to theoretical feasibility of the learning approach, as well as empirically demonstrates the efficiency: the method needs considerably fewer
function evaluations and less wall-clock time.

\subsection{Further extension and application}
\subsubsection{Extension to spatio-temporal model}
The studies from a spatio-temporal view on modeling the process are relative scarce, while the need for handling asynchronous event sequences with spatial information is increasing, in fields like criminology \cite{criminology}, epidemiology \cite{influenza} and so on. The conditional intensity function introduces new geolocation dimensions, which reads as $ \lambda^*_m(\bm{\mathrm{x}},t) = \lambda_m (\bm{\mathrm{x}},t|\mathcal{H}{(t)})$.  Here we give a brief introduction on the recently proposed spatio-temporal point process models based on deep neural network.
\begin{figure*}
    \centering
    \includegraphics[width=7.2in]{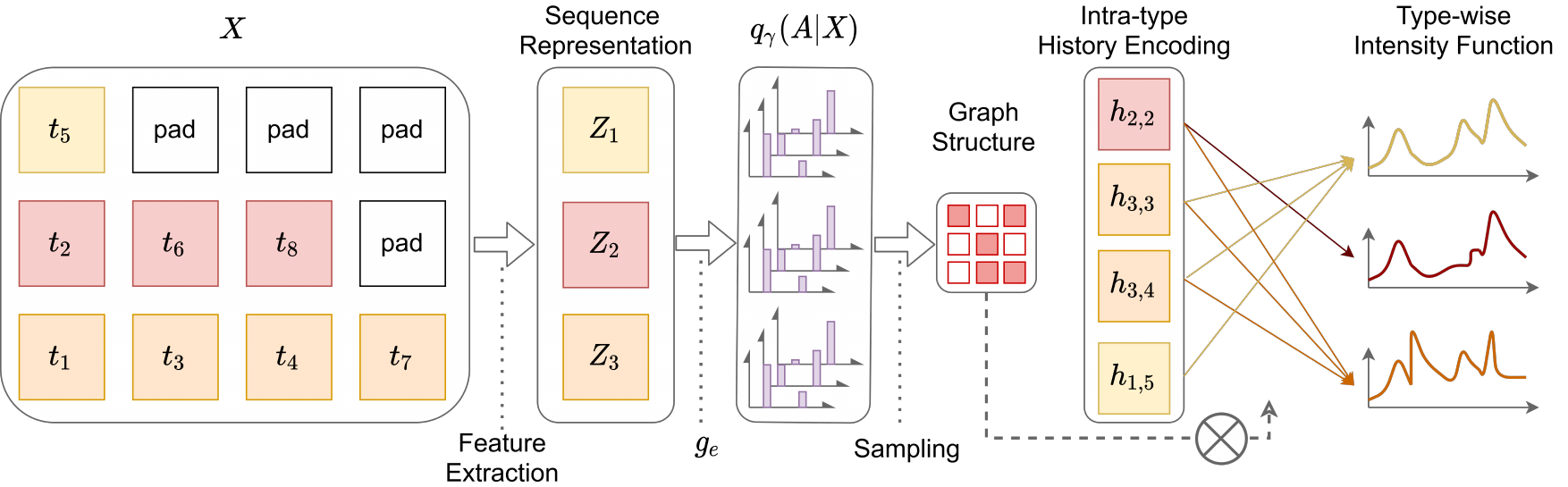}
    \caption{An illustration of the framework: The sequences of three different event types after padding are represented by $\{\bm{Z}_m\}_{1\leq m \leq 3}$. After $g_e(\cdot)$, the discrete distribution of each element in $\bm{A} \in \mathbb{R}^{3 \times 3}$ is formulated. The sampled adjacency matrix determines the message passing from intra-type history encoding to the type-wise CIF.}
    \label{figure:variationalframe}
  \end{figure*}
\\
\\
\textbf{DMPP \cite{Okawa2019}.}  Besides integrating the historical events, Deep Mixture Point Process (DMPP) also leverages contextural features $\mathcal{D}$ to fit the spatio-temporal CIF. The formulation of the CIF reads
{\small
\begin{align}
    \lambda_m (\bm{\mathrm{x}},t|\mathcal{H}({t}), \mathcal{D}) =\sum_{l=1}^Lf_m(\bm{\mathrm{x}}_{-l},t_{-l}|\mathcal{H}({t}), \mathcal{D}) 
    \mathrm{ker}(\bm{\mathrm{x}}, t, \bm{\mathrm{x}}_{-l}, t_{-l}) ,
\end{align}
}%
where $(\bm{\mathrm{{x}}}_{-l}, t_{-l})$ are the last $l$-th historical events' location and time before $t$. $L$ is a window size or lag step. The intensity is defined as a mixture of kernel experts, which is choosen as Gaussian 
{\small\begin{align*}
\mathrm{ker}(\bm{\mathrm{x}}, t, \bm{\mathrm{y}}, s) 
= \exp(-([\bm{\mathrm{x}}, t] - [\bm{\mathrm{y}}, s])^T\Sigma^{-1}([\bm{\mathrm{x}}, t] - [\bm{\mathrm{y}}, s])),
\end{align*}}%
 in which mixture weights are modeled by a deep neural
network whose inputs are contextual features $\mathcal{D}$. In this formulation, the CHF is computational manageable.
\\
\\
\textbf{NEST \cite{zhu2021imitation}.} In Neural Embedding Spatio-Temporal (NEST), a mixture of spatially heterogeneous Gaussian diffusion kernel is used to approximate the impact function, and the all past events impact are all considered, which reads
\begin{align}
    \lambda_m (\bm{\mathrm{x}},t|\mathcal{H}_{t}) &=\lambda_0 + \sum_{j:t_j<t}\nu(\bm{\mathrm{x}}, t, \bm{\mathrm{x}}_l, t_l) ;\\
    \nu(\bm{\mathrm{x}}, t, \bm{\mathrm{x}}_l, t_l) &= \sum_{k=1}^K \alpha_{\bm{\mathrm{x}}_l}^{(k)} \mathrm{ker}^{(k)}(\bm{\mathrm{x}}, t, \bm{\mathrm{x}}_l, t_l),
\end{align}
where the kernel is formulated as heterogeneous Gaussian diffusion forms,
\small
\begin{align}
    \mathrm{ker}^{(k)}(\bm{\mathrm{x}},& t, \bm{\mathrm{x}}_l, t_l)= \frac{C\exp{(-\beta (t - t_l))}}{2\pi|\mathrm{det}(\Sigma_{\bm{\mathrm{x}}_l}^{(k)})|(t-t_l)} \cdot \notag \\
    & \exp \left\{ - \frac{(\bm{\mathrm{x}} - \bm{\mathrm{x}}_l - \mu_{\bm{\mathrm{x}}_l}^{(k)})^T (\Sigma_{\bm{\mathrm{x}}_l}^{(k)})^{-1} (\bm{\mathrm{x}} - \bm{\mathrm{x}}_l - \mu_{\bm{\mathrm{x}}_l}^{(k)})} {2 (t-t_l)} \right\}.
\end{align}
\normalsize

The formulation has a closed-form integral with a negligible term which is computationally intractable. 
Different from the former works, the proposed heterogeneous Gaussian diffusion kernel in the NEST can capture a more complicated
spatial-nonhomogeneous structure.

The imitation learning approach is used, similar to the discussed reinforcement learning in Sec. \ref{sec:4.9}. 
It aims to reduce the gap between the actual divergence between the training data and the sequence generated from the model, by using policy gradient with variance reduction.
\textbf{NSTPP \cite{chen2021neural}.} In Neural Spatio-Temporal Point Process (NSTPP), the Continuous-time Normalizing Flows (CNF) which is based on Neural ODE \cite{chen2019neural} is employed as a more flexible method, with the spatio-temporal joint intensity is decompsed as 
\begin{align}
    \lambda^*_m(\bm{\mathrm{x}},t) = \lambda^*_m(t) f^*(\bm{\mathrm{x}}|t),
\end{align}
in which the spatial distribution $f^*(\bm{\mathrm{x}}|t)$ is modeled by Jump CNF, and the attention mechanisims based on the Transformer architecture is used to encode the history.
The model parameterize the process by combining ideas from Neural Jump Stochastic Differential Equations \cite{jia2020neural} and Continuous Normalizing
Flows to create highly flexible models. Besides, it still allow exact likelihood computation. Because the spatio-temporal model is not what we focus on, and preliminaries on Neural ODE are so complicated, we do not give further discription on the NSTPP.

\subsubsection{Practical applications}
 A series of works on dynamic graphs \cite{xu2020inductive, sankar2019dynamic, trivedi2018dyrep} employ the temporal point process to model the interaction between nodes or addition and removal of new nodes, such as user communication pattern and new user registration behavior in social network.
 Temporal point process with deep learning is also applied to the electronic health records \cite{enguehard2020neural}, in which important problems of multi-label setting in health records are solved. 
Besides, due to the impact of the epidemic in recent years, more and more emphasis is laid on its application to infectious disease diffusion \cite{influenza, li2021understanding, epidemiology}, aiming to capture both the spatial spread patterns and figure out the temporal dynamics of the disease.

\section{A variational framework for Granger Causality Discovery}
\subsection{A variational framework}
\begin{figure*}[!htb]
  \centering
  \includegraphics[width=6.0in]{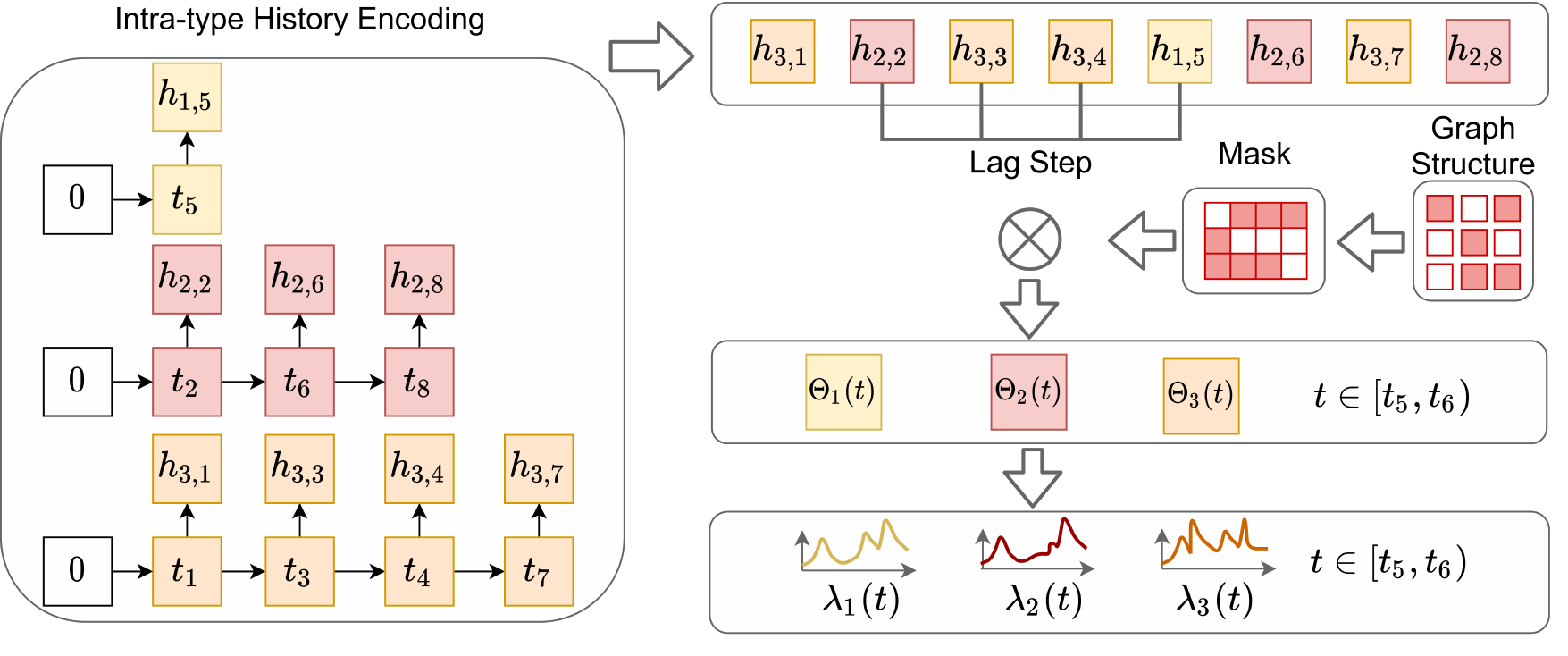}
  \caption{An example of intra-type history encoding and type-wise intensity: The whole sequence is firstly split into multivariate series according to their event types. The history encoder operates on each series to obtain intra-type history encoding. The mask generated by latent graph structure is used to govern the message passing process from intra-type history encoding to type-wise CIFs.}
  \label{figure:intratypeembed}
\end{figure*}
Because the deep-learning-based methods on relational discovery for temporal point process are quite scarce, we aim to propose a framework for Granger causality discovery in temporal point process, which is of greater interpretability than previously proposed events' relational discovery. As discussed before,  the discovery of the Granger causality can be formulated as the inference of the unknown Granger causality graph structure.
We take the adjacency matrix $\bm{A}$ of the graph as a latent discrete matrix, and employ the variational autoencoder (VAE) \citep{vae} framework as used in neural relational inference (NRI) \citep{nri} for both likelihood maximum with the decoder $q_{\theta}(\bm{X}|\bm{A})$, and graph structure inference with the encoder $p_{\gamma}(\bm{A}|\bm{X})$, where $\bm{X}$ is the input sequences of event types and timestamps. Our target is to maximize the evidence lower bound (ELBO) on all $S$ samples of sequences:
\begin{align}
\label{eq:ELBO}
    &\mathbb{E}_{p_{\gamma}(\bm{A}|\bm{X})}[\log q_{\theta}(\bm{X}|\bm{A})] - \mathrm{KL}[p_{\gamma}(\bm{A}|\bm{X})||p(\bm{A})],
\end{align}
 where $p(\bm{A})$ is the prior distribution of the adjacency matrix, leading to the KL term for a uniform prior as the sum of entropies \citep{nri}.
The term $\mathbb{E}_{p_{\gamma}(\bm{A}|\bm{X})}[\log q_{\theta}(\bm{X}|\bm{A})]$ is estimated by 
\begin{align}
\label{eq:expectation}
&\frac{1}{S}\sum_{s=1}^{S}\sum_{i=1}^{N}\log \lambda_{m^{s}_i}(t_i^{s}; \bm{\Theta}_{m^{s}_i}(t)|\mathcal{H}^{s}(t),\bm{A}^{s}) \notag \\
&- \frac{1}{S} \sum_{s=1}^{S} \sum_{m=1}^{M}\int_0^{T} \lambda_m(t; \bm{\Theta}_m(t)|\mathcal{H}^{s}(t),\bm{A}^s)dt,   
\end{align}
where $\bm{A}^s \sim p_{\gamma}(\bm{A}|\bm{X})$, and the formulation of $p_{\gamma}(\bm{A}|\bm{X})$ will be introduced in Sec. \ref{sec:seq_enc}. How $\bm{A}^s$ determines the Granger causalities among types of events will be discussed in Sec. \ref{sec:granger_frame}. Appendix C shows detailed deduction of ELBO.
\subsection{Multi-type sequence to graph structure}
\label{sec:seq_enc}
To formulate the conditional distribution $p_{\gamma}(\bm{A}|\bm{X})$ of the adjacency matrix $\bm{A}$ given the input sequences $\bm{X}$, due to the sequences of different types with unequal lengths, we first pad them with time $\max\{t_i\}$ and event type $0$, and then use \textit{1d convolution} with \textit{readout} to extract the higher-level representation $\bm{Z}_m$ of the type-$m$ sequence from the given event embeddings. Finally, we formulate the distribution of $\bm{A}_{m',m}$ as
$q_{\gamma}(\bm{A}_{m',m}| \bm{X}) = \text{sigmoid} (g_e([\bm{Z}_{m'}; \bm{Z}_{m}]))$, where the function $g_e(\cdot)$ is a linear transformation, which
breaking the symmetry of the latent graph structure.  And the $\text{sigmoid}(\cdot)$ maps the value into $[0,1]$, thus referring the probability of whether there exists a direct edge form type-$m$ to type-$m'$ . 

However, in sampling process, the reparameterization trick cannot be used to back-propagate though. One of the choices for discrete distribution optimization is to take Gumbel reparamization trick \citep{gumbel1, gumbel2}. The sample are drawn as $\bm{A}_{m',m} = \text{sigmoid}((g_e([\bm{Z}_{m'}; \bm{Z}_m]) + v) / \epsilon)$, where $v \sim \text{Gumbel}(0,1)$ and $\epsilon$ is a temperature term, leading $q_{\theta}(\bm{A}_{m',m}| \bm{X})$ to converge to one-hot categorical distribution when $\epsilon \rightarrow 0$.

\subsection{Intra-type history encoding and type-wise conditional intensity}
\label{sec:granger_frame}
Differing from previous work, because we aim to make use of the graph structure to govern the relations between types of events, the history encoding is computed in each type of events' sequence, to stop other types of events' messages from flowing into each other, and thus we call it intra-type history encoding. For an event with timestamp $t_i$ and type $m_i$, the intra-type history encoding is denoted by $\bm{h}_{m_i,i}$, with $\bm{h}_{m_i,i} = \bm{H}([\bm{e}_{m_i,1};\bm{e}_{m_i,2};\ldots;\bm{e}_{m_i,i-1}])$ comparing with Eq.~\ref{eq:histenc}. A lag step $L$ is set to determined the window size of intra-type history encoding, which is used for modeling the CIF of each event. More specifically, for type-$m$ event's CIF, the parameter is calculated by
\begin{align}
\label{eq:grangerequation}
    \bm{\Theta}_m(t) &= \bm{\chi}_m(\sum_{l=1}^L\rho_l \bm{A}_{m, m_{i-l}}\bm{h}_{m_{i-l},i-l}),
\end{align}
for $t\in[t_{i-1}, t_{i})$, where $L$ is the lag-step, $\{\rho_l\}_{1\leq l \leq L}$ are leanable weights, $\bm{A}_{m,m_{i-l}}$ is the $m$-th row, $m_{i-l}$-th column of $\bm{A}$ as a mask to permit or stop all the message passing form type $m_{i-l}$ events to type $m$. To be self-contained, it can be regarded as $\bm{h_i} = \sum_{l=1}^L\rho_l \bm{A}_{m, m_{i-l}}\bm{h}_{m_{i-l},i-l}$ in Eq. \ref{eq:intensitytrans}. And the detailed implementation of log-likelihood computation of type-wise intensities are provided in Appendix E.
\\
\begin{table*}[htb]
  \renewcommand\arraystretch{1.5}
  \centering
  \resizebox{1.7\columnwidth}{!}{
  \begin{tabular}{l|l|l}
  \toprule
  \multirow{3}{*}{History Encoder} & Recurrent Based      & RNN, LSTM, GRU                                                 \\ \cline{2-3} 
                                   & Attention Based      & Transformer                                                    \\ \cline{2-3} 
                                   & FFT Based            & FNet                                                           \\ \midrule
  \multirow{2}{*}{Overall CIF}     & FNN Based            & FNNIntegral                                                    \\ \cline{2-3} 
                                   & Mixture Distribution & Log-normal, Gompertz, Exp-decay, Weibull, Log-Cauchy, Gaussian \\ \midrule
  Type-wise CIF                    & Mixture Distribution & Log-normal, Gompertz, Exp-decay, Weibull, Log-Cauchy, Gaussian \\ \bottomrule
  \end{tabular}
  }
  \vspace*{0.2cm}
  \caption{The history encoders and conditional intensities included in empirical studies of our EDTPP. '\textit{Overall CIF}' means distribution of arrival time of different types' events are modeled in a single conditional intensity, while '\textit{Type-wise CIF}' models the distributions of each types of events.}
  \label{tab:combinations}
\end{table*}
\textit{\textbf{Proposition 1.}}
With the parameters of CIF of every type calculated by Eq. \ref{eq:grangerequation}, if $\bm{A}_{m,m'}$ as the $m$-th row, $m'$-th column of latent graph's adjacency matrix equals 0, then type $m$ does not Granger-cause type $m'$. 

The proof is provided in Appendix F. According to the proposition stated above, we claim that the inference graph by our framework both governs message passing process as well as represents the Granger causality among events. 

Our framework can be applied to all the proposed history encoders and most proposed intensity functions which is able to fit type-wise CIF. 
\\
\section{Experiments}
After dismantling and remodularizing the four parts, we first give a fair empirical study to evaluate the contributions of the two necessary parts to performance improvements. And then, we use the existing modules to solve a more difficult task as the other experiment setting -- modeling the type-wise intensities -- to illustrate the existing challenges. Finally, we evaluate our proposed Granger causality discovery framework on both real-world datasets, and reveal some problems for future research in the field of deep temporal point process according to our experiments analysis.
\subsection{Outline}
The experiments of our extensive deep temporal point process are conducted to answer the following questions:
\begin{itemize}
    \item \textbf{Contributions of the necessary parts :} With the same experiment setting, what part is of great significance in deep temporal point process? History encoders or conditional intensity?
    \item \textbf{Comparisons in different experimental settings :} There are two types of experiment setting in previous works: modeling the overall conditional intensities (\textit{Overall CIF}) which regards the distribution of the arrival time of different types' events as a single one, and modeling the type-wise conditional intensities (\textit{Type-wise CIF}) which tries to recognize the intra- and inter- patterns of each types' events. Intuitively speaking, approximating type-wise conditional intensities is more difficult than approximating the overall ones. Therefore, how big is the gap and in what aspects do the challenges of the former one exist?   
    \item \textbf{Performance of the proposed framework:} Is the proposed variational framework able to discover the Granger causalities among events without loss of fitting and predicting ability? 
    \item \textbf{Problems emerged in the experiments:} What problems still exist in the deep temporal point process according to the empirical study?
\end{itemize}
\subsection{Experiment setup}
\textbf{Methods.} We choose different combinations of history encoders and conditional intensities for a fair comparison, which is listed in Table.~\ref{tab:combinations}. 
As discussed, there are two experiment settings: Overall CIF modeling and Type-wise CIF modeling. 
Besides, it is noted that \textit{FNNIntegral} approximates the overall conditional intensity and cannot be extended to type-wise ones, so in our empirical evaluation, it is not used in the setting of type-wise intensity. Further extension of it to model type-wise intensity will be established in the future.
\\
\\
\textbf{Datasets.}  We choose two commonly-used real-world datasets: MOOC and StackOverFlow -- to evaluate different combinations of history encoders and conditional intensities.
The maximum length of the sequences is cut into 256. The detailed descriptions are shown in Table.~\ref{tab:realwolddata} and Appendix G.1.
For some of the distributions may suffer from numerical instability, we normalize the timestamps into the interval of $[0, 50.0]$ by 
    $t_{norm} = \frac{50\cdot t}{\max{t}}$, where the $\max{t}$ is obtained in training set.
\\
\begin{table}[htb]
    \centering
    \resizebox{0.98\columnwidth}{!}{
    \begin{tabular}{ccc}
    \toprule
    Statistics      &MOOC &  Stack Overflow \\
    \midrule
    Number of event types& 97 &22\\
    Numer of sequences & 7047        & 6633  \\
    Mean length &  56.28 & 72.42    \\ 
    Min length & 4&41\\
    Max length & 493&736\\
    \bottomrule
    \end{tabular}
    }
    \\
    \vspace{0.2cm}
    \caption{Raw Dataset Statistics}\vspace*{-0.2cm}
\label{tab:realwolddata}
\end{table}
\\
\begin{table*}[]\renewcommand\arraystretch{1.2}
    \resizebox{2.06\columnwidth}{!}{
    \begin{tabular}{crrrrrrrr}
    \toprule
    \multicolumn{1}{c}{\multirow{2}{*}{Methods}}                             & \multicolumn{4}{c}{MOOC}                                                                            & \multicolumn{4}{c}{Stack Overflow}                                                                 \\
                                 \cmidrule(lr){2-5} \cmidrule(lr){6-9} 
    \multicolumn{1}{c}{}                        & \multicolumn{1}{c}{NLL} & \multicolumn{1}{c}{MAPE}                    & \multicolumn{1}{c}{Top-1 ACC}               & \multicolumn{1}{c}{Top-3 ACC}                  & \multicolumn{1}{c}{NLL}                    &\multicolumn{1}{c}{MAPE}                  & \multicolumn{1}{c}{Top-1 ACC}              & \multicolumn{1}{c}{Top-3 ACC}                \\
    \midrule
    RNN+FNNIntegral        & -4.6372±0.0721          & 71.2245±4.6802          & 0.3774±0.0206          & 0.7065±0.0132          & 0.4836±0.0082           & 6.7625±0.0969          & 0.5289±0.0013          & 0.8467±0.0069          \\
    RNN+LogNormMix         & \textbf{-5.4917±0.1032} & 70.8905±1.2761          & 0.3872±0.0143          & 0.7121±0.0107          & 0.4656±0.0021           & 13.5367±2.0584         & 0.5280±0.0010          & 0.8524±0.0015          \\
    RNN+GomptMix           & -4.7599±0.0056          & \textbf{67.0907±0.5564} & 0.3993±0.0040          & 0.7177±0.0026          & 0.4682±0.0023           & 6.7642±0.0970          & 0.5293±0.0013          & 0.8501±0.0044          \\
    RNN+ExpDecayMix        & -4.0705±0.0168          & 94.3432±0.0000          & 0.3781±0.0187          & 0.7065±0.0126          & 0.4829±0.0027           & 76.2142±2.0540         & 0.5299±0.0007          & 0.8419±0.0051          \\
    RNN+WeibMix            & -4.4008±0.0176          & 75.6147±7.3159          & 0.3816±0.0123          & 0.7088±0.0084          & 0.4724±0.0014           & 7.1934±0.1334          & 0.5290±0.0024          & 0.8489±0.0058          \\
    RNN+LogCauMix          & -3.1480±0.0007          & 88.3941±0.0188          & 0.3939±0.0062          & 0.7158±0.0021          & 0.9192±0.0011           & 14.8298±0.0200         & 0.5300±0.0008          & 0.8479±0.0061          \\
    RNN+GaussianMix        & -3.0015±0.0036          & 94.2446±0.0837          & 0.3848±0.0019          & 0.7109±0.0014          & 0.4999±0.0014           & 18.6898±0.3724         & 0.5308±0.0005          & 0.8463±0.0056          \\
    \midrule
    LSTM+FNNIntegral       & -4.6458±0.0643          & 69.0532±3.9619          & 0.3903±0.0266          & 0.7124±0.0147          & 0.4652±0.0030           & \textbf{6.3803±0.0551} & 0.5284±0.0023          & 0.8612±0.0025          \\
    LSTM+LogNormMix        & \textbf{-5.4855±0.0875} & 70.4845±1.1507          & 0.3994±0.0150          & 0.7169±0.0120          & \textbf{0.4498±0.0006}  & 14.0067±2.6871         & 0.5308±0.0022          & 0.8631±0.0017          \\
    LSTM+GomptMix          & -4.7823±0.0118          & \textbf{66.2321±0.6970} & \textbf{0.4095±0.0036} & \textbf{0.7260±0.0018} & \textbf{0.4511±0.0013}  & 6.4495±0.0236          & 0.5294±0.0015          & 0.8627±0.0016          \\
    LSTM+ExpDecayMix       & -4.0854±0.0125          & 94.3417±0.0029          & 0.3994±0.0138          & 0.7175±0.0095          & 0.4610±0.0014           & 75.4951±0.0032         & 0.5309±0.0014          & 0.8636±0.0008          \\
    LSTM+WeibMix           & -4.4491±0.0338          & 71.6484±1.5004          & \textbf{0.4105±0.0387} & 0.7194±0.0116          & 0.4564±0.0013           & 7.6869±1.6173          & 0.5318±0.0011          & 0.8642±0.0005          \\
    LSTM+LogCauMix         & -3.1510±0.0002          & 88.4560±0.0145          & \textbf{0.4105±0.0017} & \textbf{0.7262±0.0014} & 0.9132±0.0003           & 14.9364±0.0320         & 0.5308±0.0014          & 0.8622±0.0018          \\
    LSTM+GaussianMix       & -3.0105±0.0017          & 94.2679±0.0624          & 0.4021±0.0017          & \textbf{0.7227±0.0010} & 0.4856±0.0013           & 17.8925±0.1867         & 0.5294±0.0031          & 0.8617±0.0021          \\
    \midrule
    GRU+FNNIntegral        & -4.6561±0.0753          & 75.0395±5.0300          & 0.3903±0.0231          & 0.7119±0.0150          & 0.4643±0.0030           & \textbf{6.4094±0.0902} & 0.5287±0.0036          & 0.8630±0.0016          \\
    GRU+LogNormMix         & \textbf{-5.5703±0.1030} & 71.5751±0.8455          & 0.3897±0.0175          & 0.7095±0.0113          & \textbf{0.4485±0.0007}  & 13.8746±1.1524         & 0.5324±0.0022          & \textbf{0.8645±0.0010} \\
    GRU+GomptMix           & -4.7798±0.0127          & \textbf{66.7481±0.2380} & \textbf{0.4097±0.0035} & \textbf{0.7255±0.0032} & \textbf{0.4507±0.0014}  & \textbf{6.4436±0.0597} & 0.5310±0.0013          & \textbf{0.8643±0.0012} \\
    GRU+ExpDecayMix        & -4.0790±0.0165          & 94.3432±0.0000          & 0.3941±0.0145          & 0.7158±0.0118          & 0.4601±0.0017           & 77.4951±1.2075         & 0.5299±0.0033          & 0.8634±0.0010          \\
    GRU+WeibMix            & -4.3979±0.0141          & 71.1910±2.7662          & \textbf{0.4099±0.0061} & \textbf{0.7249±0.0032} & 0.4551±0.0011           & 7.4052±0.6293          & 0.5318±0.0008          & \textbf{0.8645±0.0008} \\
    GRU+LogCauMix          & -3.1496±0.0004          & 88.4150±0.0212          & 0.4047±0.0036          & 0.7223±0.0025          & 0.9071±0.0099           & 15.0163±0.0928         & 0.5261±0.0089          & 0.8574±0.0135          \\
    GRU+GaussianMix        & -3.0023±0.0052          & 94.3242±0.0301          & 0.3961±0.0099          & 0.7169±0.0076          & 0.4854±0.0018           & 17.8007±0.2030         & 0.5306±0.0023          & 0.8634±0.0010          \\
    \midrule
    Attention+FNNIntegral  & -4.8006±0.0116          & \textbf{67.0141±3.1392} & 0.3718±0.0164          & 0.6973±0.0116          & 0.4594±0.0022           & \textbf{6.3298±0.1184} & 0.5347±0.0017          & 0.8632±0.0019          \\
    Attention+LogNormMix   & \textbf{-5.6623±0.0861} & 70.4523±1.1894          & 0.3801±0.0065          & 0.7050±0.0052          & \textbf{0.4486±0.0012}  & 9.9921±0.6307          & \textbf{0.5354±0.0003} & 0.8639±0.0006          \\
    Attention+GomptMix     & -4.8150±0.0019          & \textbf{68.1820±0.4076} & 0.3862±0.0032          & 0.7131±0.0010          & 0.4519±0.0010           & \textbf{6.3427±0.0757} & \textbf{0.5353±0.0004} & 0.8642±0.0006          \\
    Attention+ExpDecayMix  & -4.5714±0.0093          & 94.3432±0.0000          & 0.3496±0.0371          & 0.6814±0.0323          & 0.4610±0.0005           & 77.4925±0.0049         & \textbf{0.5347±0.0011} & 0.8633±0.0014          \\
    Attention+WeibMix      & -4.5977±0.0203          & 70.0913±1.5040          & 0.3599±0.0383          & 0.6900±0.0310          & 0.4515±0.0008           & 6.8410±0.3670          & \textbf{0.5348±0.0007} & \textbf{0.8646±0.0009} \\
    Attention+LogCauMix   & -3.1511±0.0799          & 88.4327±1.8191          & 0.3856±0.0045          & 0.7162±0.0137          & 0.8867±0.0013           & 15.1702±0.1226         & 0.5112±0.0018          & 0.8511±0.0023          \\
    Attention+GaussianMix  & -2.9915±0.0054          & 94.3397±0.0062          & 0.3517±0.0085          & 0.6840±0.0083          & 0.4874±0.0011           & 17.4988±0.4591         & \textbf{0.5351±0.0014} & \textbf{0.8646±0.0003} \\
    \midrule
    FNet+FNNIntegral       & -4.6146±0.0565          & 69.1444±3.1313          & 0.2788±0.0048          & 0.6155±0.0048          & 0.5118±0.0006           & 7.4106±0.0721          & 0.5211±0.0010          & 0.8308±0.0047          \\
    FNet+LogNormMix        & \textbf{-5.2609±0.0735} & 73.0943±0.7302          & 0.2790±0.0042          & 0.6167±0.0057          & 0.4897±0.0224           & 8.7675±1.2005          & 0.5162±0.0091          & 0.8293±0.0054          \\
    FNet+GomptMix          & -4.6773±0.0129          & 70.5275±0.6965          & 0.2811±0.0020          & 0.6176±0.0062          & 0.4945±0.0172           & 7.2800±0.1681          & 0.5146±0.0093          & 0.8259±0.0086          \\
    FNet+ExpDecayMix       & -3.9384±0.0082          & 94.3432±0.0000          & 0.2782±0.0033          & 0.6187±0.0057          & 0.5095±0.0049           & 64.5925±2.5805         & 0.5138±0.0078          & 0.8278±0.0053          \\
    FNet+WeibMix           & -4.3411±0.0043          & 70.6751±0.6197          & 0.2785±0.0026          & 0.6208±0.0040          & 0.4950±0.0197           & 27.0789±2.1900         & 0.5159±0.0104          & 0.8268±0.0102          \\
    FNet+LogCauMix         & -3.1450±0.0002          & 88.4288±0.0161          & 0.2845±0.0005          & 0.6244±0.0015          & 0.9104±0.0153           & 14.8423±0.1718         & 0.5111±0.0109          & 0.8256±0.0099          \\
    FNet+GaussianMix       & -2.9680±0.0111          & 94.2718±0.1357          & 0.2807±0.0055          & 0.6145±0.0079          & 0.5254±0.0181           & 20.2463±0.7746         & 0.5160±0.0090          & 0.8288±0.0058         
    \\\bottomrule
    \end{tabular}
}
\vspace*{0.2cm}
    \caption{Experimental results of modeling the overall CIF with different combinations of history encoder and family of distribution. The metrics in \textbf{bold} means that the model achieves the top 5 performance in the column. The mean and variance are obtained by models with the lowest five NLLs.}
    \label{tab:overallresult}
\end{table*}
\\
\textbf{Protocol.} In training process, hyper-parameters of every model of different combinations are tuned in the range of $\text{'learning rate'}:\{1\times 10^{-3}, 5\times 10^{-4}, 1\times 10^{-4}\}$, $\text{'embed dim'}:\{8,16,32\}$, $\text{'layer number'}:\{1,2,3\}$, where 'embed dim' is the dimension of history embedding, i.e. $D$. 
In the models with mixture distribution as CIF, the component distribution number is chosen as $16$, which leads that the order of parameter number is close to \textit{FNNIntegral} intensity.
And for our framework, the lag-step $L$ is set as 32.
In evaluation process, when our variational framework is used, the latent graph is set as the mean of the inferred graphs by sequences in training set. The reported metrics are the results of models chosen among all the combinations according to the lowest five negative log-likelihood (\textit{NLL}), and the hyper-parameters are tuned on validation set with early stopping technique.
We provide both mean and variance of the five \textit{NLL} to evaluate the goodness-of-fit, \textit{MAPE} to evaluate the predictive performance of next event arrival time, and \textit{Top-1 ACC} and \textit{Top-3 ACC} to evaluate the predictive performance of next event type.  The metrics are given as follows,
\begin{align}
    &\mathrm{MAPE}(\{\hat t_i\}_{1\leq i \leq N}, \{t_i\}_{1\leq i \leq N}) = \frac{1}{N}\sum_{i=1}^N\left | \frac{\hat t_i - t_{i-1}}{t_{i} - t_{i-1}} \right |;\\
    &\mathrm{ACC}_{k}(\{\hat m_i\}_{1\leq i \leq N}, \{m_i\}_{1\leq i \leq N}) \notag\\
    &\quad \quad \quad \quad = \frac{|\{m_i \in \mathrm{Top}_k\{\mathrm{logit}(\hat m_i)\} : 1\leq i\leq N \}|}{N},
\end{align}
where $\hat t_i$ is the $i$-th predicted arrival time, $\mathrm{ACC}_{k}$ is the \textit{Top-$k$ ACC}, and $\mathrm{logit}(\hat m_i) \in \mathbb{R}^M$ is obtained by Eq.~\ref{eq:typelogit} to measure the predicted discrete probability. The lower is the \textit{MAPE}, and the higher is the \text{Top-}$k$ \text{ACC}, the better performance does the model achieve.
\subsection{Extensive performance comparision}
\begin{table*}[]\renewcommand\arraystretch{1.2}
    \resizebox{2.06\columnwidth}{!}{
    \begin{tabular}{crrrrrrrr}
    \toprule
    \multicolumn{1}{c}{\multirow{2}{*}{Methods}}                             & \multicolumn{4}{c}{MOOC}                                                                            & \multicolumn{4}{c}{Stack Overflow}                                                                 \\
                                 \cmidrule(lr){2-5} \cmidrule(lr){6-9} 
    \multicolumn{1}{c}{}                        & \multicolumn{1}{c}{NLL} & \multicolumn{1}{c}{MAPE}                    & \multicolumn{1}{c}{Top-1 ACC}               & \multicolumn{1}{c}{Top-3 ACC}                  & \multicolumn{1}{c}{NLL}                    &\multicolumn{1}{c}{MAPE}                  & \multicolumn{1}{c}{Top-1 ACC}              & \multicolumn{1}{c}{Top-3 ACC}                \\
    \midrule
    GRU+LogNormMix                                & -0.8447±0.0234          & 94.3431±1.0003 & 0.3862±0.0389 & 0.7074±0.0242 & 3.4536±0.0972           & 48.6852±0.3866 & 0.5142±0.0009 & 0.8342±0.0012 \\
    GRU+GomptMix                                  & -0.3366±0.0684          & 91.2007±2.7263 & 0.3757±0.0184 & 0.7035±0.0151 & 3.4756±0.0139           & 38.2877±1.2156 & 0.5106±0.0033 & 0.8418±0.0137 \\
    GRU+WeibMix                                   & -1.4801±0.1704          & 94.2242±2.0231 & 0.3097±0.0703 & 0.6361±0.0911 & 2.4966±0.0638           & 54.8806±3.3412 & 0.5246±0.0057 & 0.8383±0.0132 \\
    \midrule
    Attention+LogNormMix                          & -1.0121±0.0292          & 94.3431±3.2234 & 0.3670±0.0142 & 0.6938±0.0136 & 3.4757±0.0127           & 69.9634±5.5834 & 0.5168±0.0106 & 0.8526±0.0075 \\
    Attention+GomptMix                            & -0.3475±0.0035          & 90.4360±0.2694 & 0.3848±0.0035 & 0.7126±0.0015 & 3.4880±0.0143           & 32.2612±5.4095 & 0.5204±0.0099& 0.8545±0.0051 \\
    Attention+WeibMix                             & -1.5436±0.2107          & 92.4614±1.2378 & 0.3338±0.0235 & 0.6896±0.0630 & 2.3814±0.0251           & 52.9625±6.0586 & 0.5315±0.0016 & 0.8532±0.0075
    \\\bottomrule
    \end{tabular}
}
\vspace*{0.2cm}
    \caption{Experimental results of modeling the type-wise CIF with different combinations of history encoder and family of distributions. }
    \label{tab:typewiseresult}
\end{table*}
\begin{table*}[]\renewcommand\arraystretch{1.2}
    \resizebox{2.06\columnwidth}{!}{
    \begin{tabular}{crrrrrrrr}
    \toprule
    \multicolumn{1}{c}{\multirow{2}{*}{Methods}}                             & \multicolumn{4}{c}{MOOC}                                                                            & \multicolumn{4}{c}{Stack Overflow}                                                                 \\
                                 \cmidrule(lr){2-5} \cmidrule(lr){6-9} 
    \multicolumn{1}{c}{}                        & \multicolumn{1}{c}{NLL} & \multicolumn{1}{c}{MAPE}                    & \multicolumn{1}{c}{Top-1 ACC}               & \multicolumn{1}{c}{Top-3 ACC}                  & \multicolumn{1}{c}{NLL}                    &\multicolumn{1}{c}{MAPE}                  & \multicolumn{1}{c}{Top-1 ACC}              & \multicolumn{1}{c}{Top-3 ACC}                \\
    \midrule
    GRU+LogNormMix                                & -1.3803±0.0497         & 94.0189±1.1080 & 0.0535±0.0059 & 0.1483±0.0136 & 2.1089±0.0023           & 47.0521±2.3972 & 0.5113±0.0004 & 0.7915±0.0014 \\
    GRU+GomptMix                                  & -0.8709±0.0280          & 78.8282±1.5319 & 0.0594±0.0033 & 0.1644±0.0072 & 2.1124±0.0024           & 20.5579±2.2540 & 0.5115±0.0004 & 0.7922±0.0004 \\
    GRU+WeibMix                                   & -1.7871±0.1388          & 93.1873±1.6187 & 0.0459±0.0150 & 0.1765±0.0231 & 1.2873±0.0537           & 34.7911±1.4928 & 0.5116±0.0006 & 0.7896±0.0017 \\
    \midrule
    Attention+LogNormMix                          & -1.3872±0.0102          & 94.0336±0.1463 & 0.0497±0.0010 & 0.1398±0.0024 & 2.1142±0.0059           & 54.8287±4.8806 & 0.5112±0.0003 & 0.7882±0.0003 \\
    Attention+GomptMix                            & -0.8754±0.0023          & 77.0698±1.3205 & 0.0500±0.0003 & 0.1401±0.0028 & 2.1150±0.0019           & 21.5009±1.1407 & 0.5112±0.0005 & 0.7908±0.0012 \\
    Attention+WeibMix                             & -1.7464±0.1435          & 94.3385±0.0093 & 0.0429±0.0059 & 0.1256±0.0107 & 1.3276±0.0097           & 27.2654±1.6573 & 0.5116±0.0001 & 0.7850±0.0016
    \\\bottomrule
    \end{tabular}
}
\vspace*{0.2cm}
    \caption{Experimental results of modeling the type-wise CIF with different history encoders and families of distributions in our variational framework. }
    \label{tab:frameworkresult}
\end{table*}
\label{sec:6.3}
We firstly give a fair empirical evaluation of different combinations of history encoders and overall conditional intensities on the real-world datasets. From the Table.~\ref{tab:overallresult}, it can be concluded that
\begin{itemize}
    \item According to the column of \textit{NLL}, which demonstrates the models' goodness-of-fit, the history encoder seems to affect the fitting ability less, because \textit{RNN-based}, \textit{Attention-based} and \textit{FNet-based} methods usually achieve similar performance when the intensity functions are the same. On the other hand, the intensity functions used for approximation matter most. From the table, \textit{LogCauMix} shows the worst fitting ability. A reason for it is that Log-Cauchy distribution is an example of a super heavy-tailed distribution, while in most real-world sequences of events, the impacts of history usually last for a short time. Besides, \textit{LogNormMix} and \textit{GomptMix} usually fits the data best, although \textit{GomptMix} is likely to suffer from the numerical instability (see Appendix B.2). Thus, we claim that the \textit{LogNormMix} is state-of-the-art method to approximate the CIF, with goodness-of-fit, closed-form likelihood, expectation and sampling and good numerical stability.
    \item According to the column of \textit{MAPE}, which demonstrates the models' predictive ability of next arrival times, the choice of intensity function is also the key. \textit{FNNIntegral}, \textit{GomptMix} and \textit{WeibMix} usually predict well. However, the predictive performance is all very bad, and thus it raises a further \textbf{problem}: How to improve the performance of next-event-time prediction task? Fitting the events well does not mean predicting the next arrival time well.     
    \item According to the columns of \textit{Top-1 ACC} and \textit{Top-3 ACC}, which demonstrate the models' predictive ability of next event types, the history encoder is paramount to this, because the prediction totally depends on the history encodings. We conclude that the \textit{Attention-based} encoders usually show good predictive performance, because its capability of capturing the very long-term messages from history events. Besides, \textit{GRU} and \textit{LSTM} also perform well in MOOC dataset because the lengths of the event sequences are comparably short.   
\end{itemize}
In addition, we also claim that the \textit{NLL} and \textit{MAPE} calculated by timestamps are influenced more by the formulation of intensity, and short-term impacts which can be both captured by the five history encoders are enough to model the dynamics of arrival time, because the differences of the evaluated performance are minor among them.
In contrast, to model the dynamics of next event types, the long-term impacts from history events may contribute more, compared with the dynamics of arrival times, so the \textit{Attention-based} encoders usually outperform the others according to \textit{Top-1 ACC} and \textit{Top-3 ACC}, especially in long sequences.

In conclusion, in terms of the history encoders, \textit{GRU} and \textit{LSTM} perform well, while \textit{GRU} costs less computational resources. Transformers with \textit{Attention} mechanisims shows great expressivity.  In terms of intensity functions, \textit{LogNormMix} has many advantages over others, while \textit{GomptMix}, \textit{WeibMix} and \textit{FNNIntegral} are also good alternatives.

\subsection{Modeling type-wise intensities}
Another experiment setting is to model events' conditional intensity of each type, which is more challenging than the former task. First, if the model can accurately predict the next-event-time according to each types' intensities, the order of predicted arrival event types will be obvious. Secondly, to find the relation between types, modeling the type-wise intensity as a multi-variate time series problem is necessary, and modeling the multi-variate series is always more complex than modeling the uni-variate ones attributing to it requires recognizing patterns of both inter- and intra-sequence.
Here we use the discussed mixture distributions to evaluate the model's performance in this setting.

According to the Table.~\ref{tab:overallresult}, we choose the most representative and well-performed RNN-based history encoder -- \textit{GRU}, and Attention-based encoder, with intensities of \textit{LogNormMix}, \textit{GomptMix} and \textit{WeibMix} which show good fitting and predictive ability in time to evaluate the deep temporal point processes' performance on the task of modeling type-wise intensities.
We can conclude from Table.~\ref{tab:typewiseresult} that
\begin{itemize}
    \item According to the column of \textit{NLL}, the fitting performance of all the evaluated models in this setting shows an overall decrease. \textit{WeibMix} shows very flexible fitting performance. Besides, it is noted in MOOC dataset, because the joint loss function contains both \textit{NLL} for fitting event times and \textit{Cross Entropy (CE)} for predicting event types, it is likely that the two terms conflict each other -- when \textit{NLL} is small, \textit{Top-1 ACC} and \textit{Top-3 ACC} will also decrease. 
    \item According to the column of \textit{MAPE}, the predictive performance of type-wise event time shows a drastic decrease, compared with the performance of event time prediction in Table.~\ref{tab:overallresult}. 
    \item According to the final two columns, the predictive performance of next event type is comparable with minor differences. 
\end{itemize} 
Through this part of empirical study, we demonstrate the wide gap of difficulties between modeling type-wise intensities and overall intensities, and thus raise another \textbf{problem} for future research: How can deep temporal point process narrow the gap and improve the fitting and predictive performance in the setting of type-wise intensity, as it is unavoidable to model the inter- and intra-series patterns in multi-variate series modeling. 

\subsection{Evaluation of the proposed framework}
Different from previous works where the history encoding is obtained by inter-events encoders, in our variational inference framework the historical events are encoded intra-types to model the type-wise intensities as well as discover the relations. 
Here we try to figure out that if our framework still performs well with the same history encoders and intensities. The experiment setting is to model the type-wise intensities, so we compare the metrics with Table.~\ref{tab:typewiseresult}.
From Table.~\ref{tab:frameworkresult}, we can conclude that
\begin{itemize}
    \item In fitting and predictive tasks where event time contributes more, the performance improves compared with the same setting in Table.~\ref{tab:typewiseresult}. One of the leading reasons is that the generated graph structure provides the option to omit non-contributing event types so as to avoid the disturbance when the model is fitting a single type's intensity function. In this way, the learned history encoding is a more accurate measurement of contributions of the past to the present.
    \item However, in evaluation of next-type-prediction, obvious reduction shows in the \textit{Top-1 ACC} and \textit{Top-3 ACC}, significantly in the MOOC dataset. We boil down the reason to two points: (1) As we claimed in Sec.~\ref{sec:6.3}, the good performance of next events prediction requires encoders' capturing long-term messages from the past, but the lag-step existing in the computation of history encodings in Eq.~\ref{eq:grangerequation} unavoidably limits the messages to flow from long past events, so the overall predictive performance of next event type decreases. (2) Furthermore, the event number of MOOC is 97, more than the lag-step as a hyper-parameter which is set as 32. Therefore, some past event types with the greatest impacts may be omitted in that the number of event types outnumbers lag-step size. In Stack Overflow dataset with 22 event types, the decrease in performance is alleviated.
\end{itemize}
\begin{figure}
    \includegraphics[width=1.0\linewidth]{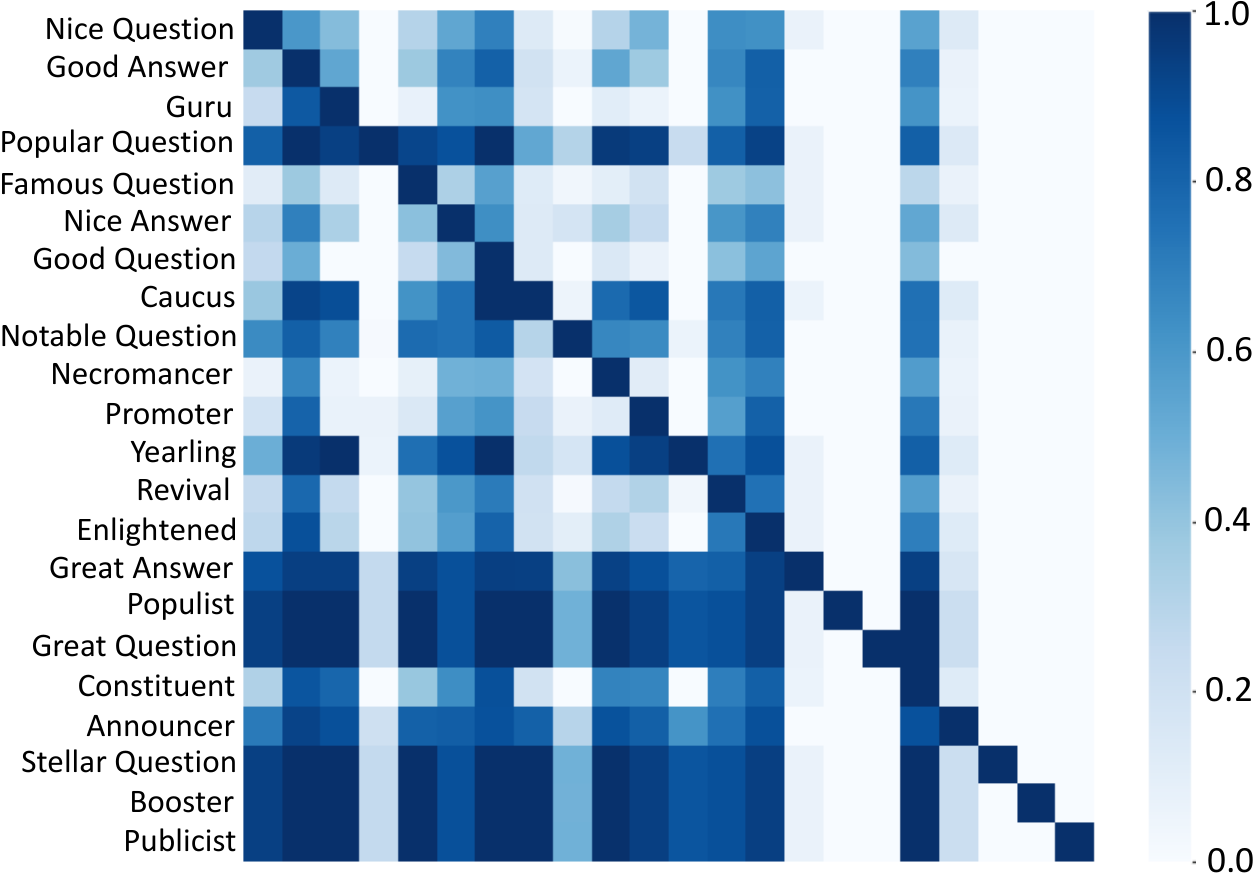}\vspace{-0.1cm}
    \caption{Visualization of the learned latent Granger causality graph. The horizonal axis as types of events has the same order as the vertical axis. } \label{fig:learnedgraph}
\end{figure} 
In this way, we conclude that our framework works well in both fitting sequences, predicting arrival times and discovering the latent graph structures, while \textbf{problems} still exist in limitation of capturing the long-term patterns and restrictions on modeling temporal point process with a large number of event types. 
Finally, the visualization of the learned graph obtained by best model of \textit{Attention+WeibMix} is given in Fig.~\ref{fig:learnedgraph}, with high sparsity compared with the learned lantent graphs in \cite{dgnpp,sahp}.

\subsection{Existing problems}
\label{sec:6.6}
At the end of this section, we list the existing problems through our empirical study, and point out some further research topics for  deep temporal point process.
\begin{itemize}
    \item \textbf{How to improve the predictive performance on next event arrival time?} Even if the chosen family of distributions can fit the intensity functions well, the predictive performance is still unsatisfactory.
    \item \textbf{How to narrow the gap between modeling type-wise intensities and overall ones?} The former one is more challenging but unavoidably necessary. Most existing methods focus on the latter one, while further research should also lay emphasis on the multi-variate settings, because the type-wise intensities can provide more information on both inter- and intra-dynamics of different types of events.
    \item \textbf{How to capture the long-term dependencies in our variational framework?} Due to the window or lag-step in our framework in aggregating messages from history, the long-term dependencies are likely to be omitted. Future works should enable our framework to capture the long-term patterns as well as discover the Granger causalities.
\end{itemize}

\section{Discussion}
Despite the recent success achieved by deep temporal point process, challenging problems still remain unsolved. In this section, we conduct analysis on other problems besides what we have discussed in Sec.~\ref{sec:6.6}, and point out promising directions for future works.
\\
\\
\textbf{Theoretical foundation.} The theoretical foundation is a common problem existing in deep learning. In history encoder's expressivity, RNN is proved theoretically with universal approximation \cite{rnnuniversal}, and some works also discuss the theorem on universal approximation of the intensity functions \cite{unipoint,lognorm,huangnaf,jaini2019sumofsquares}. However, further theoretical contributions should be made to problems like the predictive deviation which gives further insights in the expectation of next event times and the truth, and approximation ability, i.e. the relations between neuron size and the upper bound of target intensity and distributions for approximation. Besides, in the establishment of new learning approaches, the assurance of convergence and the converging speed of the iteration are also of great importance. 
\\
\\
\textbf{Explanability and interpretability.} The works on discovering the latent relations between different types of events are still scarce. Few works try to explain the mechanisms in the deep temporal point process, such as measuring the impacts from past events or disentangling the relations between different types.
In classical statistics, modeling the process should both concentrate on accuracy and interpretability, while the deep temporal point process abandons the latter one to some degree in exchange for improved accuracy. 
A series of works on explanability and interpretability from a perspective of graph neural networks have been proposed \cite{explainer1,explainer2,explainer3,explainer4,explainer5}, and some of the core ideas can be transferred into deep temporal point process.
\\
\\
\textbf{Experiment agreements.} As discussed, some of the methods are established for type-wise intensities modeling \cite{sahp}, while others focus on the overall ones. Thus, the comparison of different metrics is not meaningful due to the disagreement of experimental settings. And we appeal that the future works can all conduct experiments on both settings for performance comparison. 
Besides, more ablation studies should be conducted to highlight the contributions of the proposed methods. For example, when the contributions of one's work focus on history encoders, the intensity function should be fix, or using different intensity functions to show the overall improvements brought about by the history encoders.
Another problem is that some of the intensity functions are computationally intractable, and thus the numerical or Monte Carlo integration methods should be taken. As the time and space complexity increases dramatically, the complexity analysis is very necessary for readers to figure out if the additional burden is bearable. 
\\
\\
\textbf{Real-world datasets for evaluation.} The samples used to evaluate the model's performance are assumed to be i.i.d from some unknown generating process. However, in real-world datasets, samples are usually drawn from human behaviors like social networks. Chances are that the contextual correlation exists in the sampling. For example, the sequences sampled yesterday may now influence the user's behaviors a lot. In this scenario, the assumption is not reasonable, so the launch of suitable datasets for model evaluation and further test on the i.i.d. assumption are urged.
Besides, the data quality is also hard to guarantee. We give a visualization of histogram on the first event timestamps of different samples in MOOC and Stack Overflow. It demonstrates extremely high variance in MOOC's first event timestamps. So we doubt that if the samples are generated in identical distribution? And as no history observations are given, how can the model accurately predict the first event's arrival time when it differs widely in different samples? In this way, we call on that meticulous data analysis should be conducted on each dataset, and the quality of datasets for experimental evaluation still needs to be improved.

\begin{figure}
    \centering
            \subfigure[ MOOC.]{ 
                \includegraphics[width=0.75\linewidth]{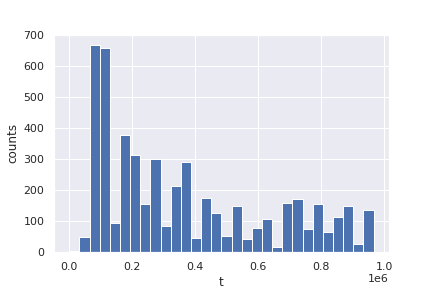}}\vspace{-0.3cm}
            \subfigure[Stack Overflow.]{
                \includegraphics[width=0.75\linewidth]{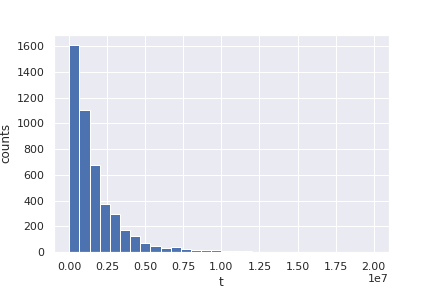}}
        \caption{First event arrival time of samples in different datasets.}
        \vspace{-0.3cm}
\end{figure} 
\section{Conclusion}
In this paper, we first summarize four key research components in deep temporal point process: encoding of history sequence, formulation of conditional intensity function, relational discovery of events and learning approaches for optimization. And then we dismantle, extend and remodularize the necessary two parts: history encoders and intensity functions for further fair empirical study.
A comprehensive case study is conducted, including most of recently proposed methods on deep temporal point process, in which these methods are also dismantled into the four parts and analyzed one by one.
Besides, a variational framework is proposed for Granger causality discovery. In the experimental section, we give a fair empirical study in two settings, and conclude three existing problems according to the results. Finally, we point out some technical limitations of the current research and provide promising directions for future work on EDTPP.
Our source code of extensive deep temporal point process (EDTPP) is available on \url{https://github.com/BIRD-TAO/EDTPP} .
\bibliographystyle{IEEEtran}
\bibliography{edtpp}

\renewcommand\thefigure{A\arabic{figure}}
\renewcommand\thetable{A\arabic{table}}
\setcounter{table}{0}
\setcounter{figure}{0}
\clearpage
\appendix
\section*{A. Notation and Preliminary of Point Process}
\label{app:A}
\begin{table}[ht]
\centering
\resizebox{1.05\columnwidth}{!}{
\begin{tabular}{lp{0.73\columnwidth}}
\toprule
Symbol                                                                                                                                  & Used for      \\    
\midrule

$t$                                                                                                                                     & Timestamp.                                                                                 \\
$m$                                                                                                                                     & Maker of type of event.                                                                    \\
$M$                                                                                                                                     & Number of types of events.                                                                 \\
$N$                                                                                                                                     & Number of events in a sequence.                                                            \\
$\lambda^*_m(t)$                                                                                                                        & Conditional intensity function of type-$m$ event at time $t$.                              \\
$\Lambda_m^*(t)$                                                                                                                          & Integrated conditional intensity function of type-$m$ event from $0$ to $t$.               \\
$\mathcal{H}(t)$                                                                                                                        & Historial event set happening before $t$.                                                  \\
$\mathcal{H}_m(t)$                                                                                                                        & Historial event of type $m$ set happening before $t$.                                                  \\
$\mathcal{N}_m(t)$                                                                                                                                &  The number of events of type $m$ occurring in the interval $[0,t)$.          \\
$f^*_m(t)$                                                                                                                             & Probalibility density function of type-$m$ event.                                          \\
$F^*_m(t)$                                                                                                                              & Cumulative distribution function of type-$m$ event.                                        \\
$\bm{\omega}(\tau)$                                                                                                                     & Transformation mapping time interval into high-dimensional space.                          \\
$\bm{e}_j$                                                                                                                              & The $j$-th event's embedding.                                                              \\
$\bm{h}_i$                                                                                                                              & The $i$-th event's history embedding.                                                      \\
$[\cdot ; \cdot]$                                                                                                                       & Concatenation of two vectors/scalars.                                                      \\
$\bm{E}$                                                                                                                                & Embedding matrix of event types.                                                           \\
$D$                                                                                                                                     & Dimension of the embedding vectors.                                                        \\
$\bm{\chi}_m(\bm{h_i})$                                                                                                                 & Mapping of type-$m$ event the history embedding into parameter space.       \\
$\Theta_m(t)$                                                                                                                           & Parameters in the type-$m$ events' conditional intensity function.                         \\
$\mathcal{G} = (\bm{V}, \bm{\mathcal{E}}, \bm{A})$ & Granger causality graph with vertex set, edge set and adjacency matrix.       \\
$\bm{A}_{m',m}$                                                                                                                         & $m'$-th row, $m$-th column of the adjacency matrix.                                        \\
$\text{RNN}(\bm{e}, \bm{h})$                                                                                                            & Recurrent neural network as history encoder.                                               \\
$\phi(\bm{e}_j, \bm{e}_i)$,$\psi(\bm{e}_j)$                                                                                             & Attention weight from $\bm{e}_j$ to $\bm{e}_i$ and value used in Attention.                \\
$\text{FFT}(\bm{e})$                                                                                                                    & Fast Fourier Transform.                                                                    \\
$\text{Top}_k\{\cdot\}$                                                                                                                 & Function to range and choose the highest $k$ values in the set.                                        \\
$w_k$                                                                                                                               & Mixture weights.                                                                           \\
$\mu_k, \sigma_k, \alpha_k, \beta_k, \eta_k$                                                                             & Parameters in different mixture models.                                                    \\
$\bm{X}$                                                                                                                                & Input sequences.                                                                           \\
$\bm{A}$                                                                                                                                & Adjacency matrix of latent graph.                                                          \\
$q_{\theta}(\bm{X}|\bm{A})$                                                                                                             & Decoder of input sequences, given graph structure.        \\
$p_{\gamma}(\bm{A}|\bm{X})$                                                                                                             & Encoder of lantent graph structure, given input sequences.                                 \\
$S$                                   & Total number of the observed sequences of events for training.\\
$\bm{Z}_m$                                                                                                                              & Higher-level representation of type-$m$ sequence.                                          \\
$\nu$                                                                                                                                   & Samples whose distribution is Gumbel.                                                      \\
$\epsilon$                                                                                                                              & Temperature term in Gumbel distribution.                                                   \\
$\bm{h}_{m_{i},i}$                                                                                                                      & Intra-type history embedding of event with timestamp $t_i$ and type $m_i$.                 \\
$L$                                                                                                                                     & Lag-step for history embedding to aggregate information from intra-type history embedding. \\
$\rho_l$                                                                                                                                & Weight of $l$-th lag intra-type history embedding aggregating to overall history embedding.   \\
\bottomrule
\end{tabular}
}
\vspace*{0.2cm}
\caption{Glossary of Notations used in this paper except in the Sec.~\ref{sec:4}.}
\end{table}
\textbf{Temporal point process with markers.} For a temporal point process $\{t_i\}_{i\geq1}$ as a real-valued stochastic process indexed on $\mathbb{N}^+$ such that $T_i\leq T_{i+1}$ almost surely (here $T_i$ representing the random variable), each
random variable is generally viewed as the arrival timestamp of an event. When each timestamp is given a type marker, i.e. $\{(t_i, m_i)\}_{i\geq1}$, the process is called marked temporal point process, also called multivariate point process as well. For a marked temporal point process, it is a coupling of $M$-dimensional point/counting process $\{\mathcal{N}_1,\mathcal{N}_2,\ldots,\mathcal{N}_M\}$. 

\noindent\textbf{Conditional intensity function and probability density function.} As defined in Eq.~\ref{eq:intensitydefine}, the Eq.~\ref{eq:cifandcdf} can be obtained through
\begin{align*}
    \lambda (t|\mathcal{H}(t))dt  &=  \mathbb{E}[\mathcal{N}(t + dt) - \mathcal{N}(t)| \mathcal{H}(t))\\
    &= \mathbb{P}(t_i \in [t, t+dt)| \mathcal{H}(t))\\
    &= \mathbb{P}(t_i \in [t, t+dt)| t_i \notin [t_{i-1},t),  \mathcal{H}(t_{i-1}))\\
    &= \frac{\mathbb{P}(t_i \in [t, t+dt), t_i \notin [t_{i-1},t)|  \mathcal{H}(t_{i-1}))}{\mathbb{P}( t_i \notin [t_{i-1},t)|  \mathcal{H}(t_{i-1})))}\\
    &= \frac{\mathbb{P}(t_i \in [t, t+dt)|  \mathcal{H}(t_{i-1}))}{\mathbb{P}( t_i \notin [t_{i-1},t)|  \mathcal{H}(t_{i-1}))}\\
    &= \frac{f(t|\mathcal{H}(t_{i-1}))}{1-F(t|\mathcal{H}(t_{i-1}))}\\
    &= \frac{f^*(t)}{1-F^*(t)},
\end{align*}
where the lower script $m$ is omitted.
In this way, the reverse relation can be given by
\begin{align*}
    f^*(t) &= \lambda^*(t)\exp(-\int_{t_{i-1}}^{t}\lambda^*(\tau)d\tau);\\
    F^*(t) &= 1- \exp(-\int_{t_{i-1}}^{t}\lambda^*(\tau)d\tau).
\end{align*}
\\
\\
\textbf{Examples of temporal point process.} Here we give some classical temporal point process examples in brief.
\\
\textbf{Example 1.} (Possion process) The (homogeneous) Poisson process is quite simply the point process where the conditional intensity function is independent of the past. For example, $\lambda^*(t) = \lambda(t) = c$ which is a constant.
\\
\textbf{Example 2.} (Hawkes process) The conditional intensity function of which can be written as 
\begin{align*}
    \lambda^*(t) = \alpha + \sum_{t_j<t} g(t-t_j;\eta_j, \beta_j), \label{example:2}
\end{align*}
which measures all the impacts of all the historical events on the target timestamp $t$. The classical Hawkes process formulates the impact function $g(t-t_j;\eta, \beta) = \eta\exp(\beta(t-t_j))$ as the exponential-decayed function. Inspired by this, we extend the family of distribution of temporal point process with so-called Exp-decayed Mixture.
\\
\textbf{Example 3.} (Self-correcting process)  While the impacts are cumulated in Hawkes process, and events are more likely to clustered in a small time interval, self-correcting point process aims to model the process that the intensity increases as time passes, by formulating the CIF as
\begin{align*}
     \lambda^*(t) = \exp(\mu t - \sum_{t_j < t}\alpha_j).
\end{align*}
Thus the chance of new points decreases immediately after a point has appeared.

\section*{B. Detailed Implementation of CIF}
\begin{Theorem}{\rm [Universal Approximation Theorem of Mixture (Theorem 33.2 in \cite{Asymptotictheory}).]} Let $p(x)$ be a continuous density on $\mathbb{R}$. If $q(x)$ is any density on $\mathbb{R}$ and is also continous, then given $\epsilon >0$, and a compact set $\mathcal{S} \in \mathbb{R}$, there exist number
of components $K \in \mathbb{N}$,  mixture coefficients $\bm{w} \in \Delta^{K-1}$, locations $\bm{\mu} \in \mathbb{R}^{K}$, and scales $\bm{s} \in \mathbb{R}_+^{K}$, s.t. for the mixture distrubution $ \hat p(x) = \sum _{k=1}^K w_k \frac{1}{s_k} q( \frac{x - \mu_k }{s_k})$, it holds $\sup_{x\in\mathcal{S}}|p(x) - \hat p(x)| < \epsilon$.
\end{Theorem}
By the theorem, we consider that use the mixture form of classical survival distribution including Weibull, Gompertz, Log-normal \citep{lognorm} and Log-Cauchy to approximate the target PDF. Besides, the Exp-decay is also included, which will be discussed latter.

\subsection*{B.1. Implementation of Log-norm Mixture}
\label{app:B1}
The implementation of Log-norm mixture is mostly based on \cite{lognorm}, which provided a stable version of a log-normal distribution. The log-normal is equivalent to following
\begin{align*}
    z_m &\sim \text{GaussianMixture}(\bm{w}_m, \bm{\mu}_m, \bm{\sigma}_m);\\
    \tau_m & = \exp(z_m),
\end{align*}
with $\bm{w}_m \in \mathbb{R}_+^{K}, \bm{\mu}_m\in \mathbb{R}^{K}, \bm{\sigma}_m \in \mathbb{R}_+^{K}$, for any $m$. In type-wise intensity framework, $\bm{\chi}(\bm{h}_i) = \bm{W}_\theta \bm{h}_i + \bm{b}_\theta$ and $\bm{W}_\theta \in \mathbb{R}^{3MK \times D}$, $\bm{b}_\theta \in \mathbb{R}^{3MK}$. Therefore,  $\bm{\chi}(\bm{h}_i)\in \mathbb{R}^{3MK}$, and $\bm{\chi}(\bm{h}_i) = [\bm{w}_1; \ldots; \bm{\mu}_M;\bm{\mu}_1;  \ldots; \bm{\mu}_M;\bm{\sigma}_1; \ldots; \bm{\sigma}_M;]$. To normalize the $\bm{w}_m$, a softmax will be stacked after, and an exp map will be added after $\bm{\sigma}_m$ to ensure it is positive. 
Mixture PDF is given in \ref{eq:cdflognorm}, and the CDF is given by
\begin{align*}
    F^*(t) = \frac{1}{2} + \sum_{k=1}^K \frac{w_k}{2} \text{erf}(\frac{\ln(t-t_{i-1}) - \mu_k}{\sqrt{2}
    \sigma_k}) \quad\quad t\in[t_{i-1}, t_{i}) .
\end{align*}
Although the CDF has no closed form, the accurate approximation of erf function can give a stable numerical value, which also permits backward propagation.
The expectation has closed form, as $ \sum_{k=1}^Kw_k\exp(\mu_k + \frac{\sigma_k^2}{2})$.

\subsection*{B.2. Implementation of Gompertz Mixture}
\label{app:B2}
In Gompertz distribution, the CIF reads $\lambda(t) = \eta \exp (\beta t)$, where $\eta, \beta > 0$. By Eq.~\ref{eq:cifandcdf}, the PDF and corresponding CDF can be obtained by
\begin{align*}
    f(t) & = \eta \exp(\beta t - \frac{\eta}{\beta} \exp(\beta t) + \frac{\eta}{\beta});\\
    F(t) &= 1 - \exp(- \frac{\eta}{\beta}\exp(\beta t) + \frac{\eta}{\beta}); \\
    \Lambda(t) &= \exp(- \frac{\eta}{\beta}\exp(\beta t) + \frac{\eta}{\beta}).
\end{align*}
Therefore, we write the mixture of Gompertz for model the process as Eq.~\ref{eq:gomppdf}. The sampling methods as follows
\begin{align*}
    k_m &\sim \text{Multinomial}(w_1, w_2, \ldots, w_K);\\
    u &\sim U[0,1];\\
    g &= \frac{1}{\beta_{k_m}}\ln(1 - \frac{\beta_{k_m}}{\eta_{k_m}}\ln(1-u)).
\end{align*}
It is easy to prove that $g \sim G(\eta_{k_m}, \beta_{k_m})$ for its CDF has inverse form.
In implementation, the numerical instability is very series, because
\begin{itemize}
    \item the $\frac{1}{\beta}$ term may goes infinity, and $- \frac{\eta}{\beta}\exp(\beta t) \rightarrow - \infty$, $ \frac{\eta}{\beta}\rightarrow + \infty$, which will cause $\Lambda(t) \rightarrow \infty$.
    \item the ${\beta}$ term may goes infinity, and $\exp(\beta t) \rightarrow +\infty$, which will cause $\lambda(t) \rightarrow \infty$.
\end{itemize}
To solve it, we clamp the value of $\beta$ in $[1\times 10^{-7}, 1\times 10^{7}]$ and clamp $\exp(\beta t_{\text{max}}) < 50$. The three parameter in the component distribution is obtained as the formerly discussed Log-norm Mixture does. \\
The renormalized technique in our experiments to restrict time interval into $[0, 50.0]$ is also to guarantee the stability of the likelihood computation in Gompertz, because some of the time intervals may event be greater than $1e7$, will will cause explosion on $\exp(\beta t)$ terms.
For prediction, we use 1-d trapezoidal integration method to approximate the expectation because the closed form does not exist.
\subsection*{B.3. Implementation of Exp-decay Mixture}
\label{app:B3}
Inspired by the impact function of classical Hawkes process formulates in Example 2, we consider a three parameter distribution, whose PDF is govern by the CIF as 
\begin{align*}
    \lambda(t) = \eta \exp(-\beta t) + \alpha,
\end{align*}
with $\eta, \beta >0$ and $\alpha \leq 0$. The intensity is decayed within exponential speed, and thus we call it Exp-decay distribution. The corresponding CDF and PDF can be obtained by 
\begin{align*}
     f(t) &= (\eta  \exp(-\beta t) + \alpha) \exp((\frac{\eta}{\beta} - 1)\exp(-\beta t) - \alpha t);\\
     F(t) &=1 - \exp((\frac{\eta}{\beta} - 1)\exp(-\beta t) - \alpha t).
\end{align*}
The mixture of Exp-decayed model has four parameters, so we need to expand the weight $\bm{W}_\theta \ \in \mathbb{R}^{4MK\times D}$ and bias $\bm{b}_\theta\in \mathbb{R}^{4MK}$. The numerical instability problem is also solved by clamp the parameter $\beta$, while it is more stabile than Gompertz distribution because the negative value on the power of exp. The expectation is also computed with numerical methods.
\subsection*{B.4. Implementation of Weibull Mixture}
\label{app:B4}
Weibull distribution is very classical in modelling the survival function, due to its great approximating ability, i.e. pdf with different parameters has different shape. The CIF, CDF, PDF of Weibull distribution is given by
\begin{align*}
    \lambda(t) &= \eta\beta(\eta t)^{\beta -1};\\
    f(t) &= \eta \beta (\eta t) ^{\beta -1} \exp{(-(\eta t)^\beta)};\\
    F(t) &= 1 - \exp(-(\eta t)^\beta),
\end{align*}
where $\beta, \eta > 0$, CIF is decreasing for $\beta < 1$, increasing for $\beta > 1$ and constant for $\beta = 1$, in which case the Weibull distribution reduces to an exponential distribution. 
There are no numerical instability in Weibull Mixture, so the parameter range is not limited to a certain range, and thus the model shows a good expressivity in the experiments. To calculate the expectation, the distribution has a closed form, as $\sum_{k=1}^{K}w_k\frac{\Gamma(1 + 1/\beta_k)}{\eta_k}$, where $\Gamma(\cdot)$ is the Gamma function.
\subsection*{B.5. Implementation of Log-Cauchy Mixture}
\label{app:B5}
The Log-Cauchy distribution is an example of a heavy-tailed distribution, with no a defined mean or standard deviation, CDF and PDF of which reads
\begin{align*}
    f(t) &= \frac{1}{t\pi} \frac{\sigma}{(\ln t - \mu)^2 + \sigma^2};\\
    F(t) &= \frac{1}{2} + \frac{1}{\pi} \arctan(\frac{\ln t - \mu}{\sigma}),
\end{align*}
the CIF of which decreases at the beginning and at the end of the distribution. The expectation is expected in a finite interval, because in theory, it does not exist.

\section*{C. Deduction of ELBO}
\label{app:C}
The ELBO in Eq.~\ref{eq:ELBO} consists of two terms. For the second $KL$-divergence term, when there is no prior knowledge, $p(\bm{A}_{i,j}=1) = \frac{1}{2}$, and
\begin{align*}
    &KL[p_{\gamma}(\bm{A}|\bm{X})||p(\bm{A})]\\ =& \sum_{i,j}\sum_{a=0,1}p_{\gamma}(\bm{A}_{i,j}=a|\bm{X})\log(p_{\gamma}(\bm{A}_{i,j}=a|\bm{X})/p(\bm{A}_{i,j}=a))\\
    =&\sum_{i,j}\sum_{a=0,1}p_{\gamma}(\bm{A}_{i,j}=a|\bm{X})\log p_{\gamma}(\bm{A}_{i,j}=a|\bm{X}) \\
    &\quad \quad \quad - c\sum_{i,j}\sum_{a=0,1}p_{\gamma}(\bm{A}_{i,j}=a|\bm{X})
\end{align*}
For the first term, the expectation is estimated by samples, i.e.
\begin{align*}
    &\mathbb{E}_{p_{\gamma}(\bm{A}|\bm{X})}[\log q_{\theta}(\bm{X}|\bm{A})]    =\frac{1}{S}\sum_{s=1}^{S}\log q_{\theta}(\bm{X^s}|\bm{A^s}),
\end{align*}
$\bm{A^s}\sim p_{\gamma}(\bm{A}|\bm{X})$, and $\bm{X}^s$ is the $s$-th input sequence, $\bm{X}^s = \{(t^s_i, m^s_i)\}_{0\leq i\leq N_s}$. $\log q_{\theta}(\bm{X^s}|\bm{A^s}) = l(\bm{\Theta}|\bm{A}^s)$. Therefore, Eq.~\ref{eq:expectation} as the estimation of the expectation term in ELBO is obtained. 

\section*{D. Difference from NRI and DGNPP}
\label{app:D}
Although the graph inference encoder is similar to NRI \cite{nri}, differences still exist. NRI is usually to handle synchronous time series data with equal length and interval, while the point process observation is asynchronous, thus requiring well-designed techniques including event embedding, sequence to vector module and batch processing. Secondly, the latent graph $\bm{A}$ are of different use. The generated adjacency matrix of Granger causality graph as a random matrix $\bm{A}$ is used to govern the message passing process which serves as allowing or stopping all the message passing process from one type of event to another in modelling the CIF,  while NRI stacks the different adjacency matrix representing the different edge types, which will probably cause a complete graph when the overall relation is regarded as the weighted sum.

Besides, the relations between types from DGNPP \cite{dgnpp} are obtained by the product of type embeddings. And it is only used in \textit{Attention-based} encoder, but ours can be implemented in all the discussed history encoders.
\section*{E. Detailed Implementation of Type-wise log-likelihood}
To maximize the likelihood, there are two terms for each type. One is the conditional intensity term, the second is its integral in the interval from the start time to the end. For the first term, we first calculate all the intensity on the observed occurrence time regardless of which type of event it is, and then use a mask which is one-hot encoding of types to set the contribution to this term from unrelated types of event zero. For the second term, the integral is computed on the whole interval, and thus once an event happens whatever its type, all the intensity functions should update to consider its impact. Therefore,  the computation of type-wise likelihood should be conduct as follows:
\begin{figure}[htb]
  \centering
  \includegraphics[width=3.5in]{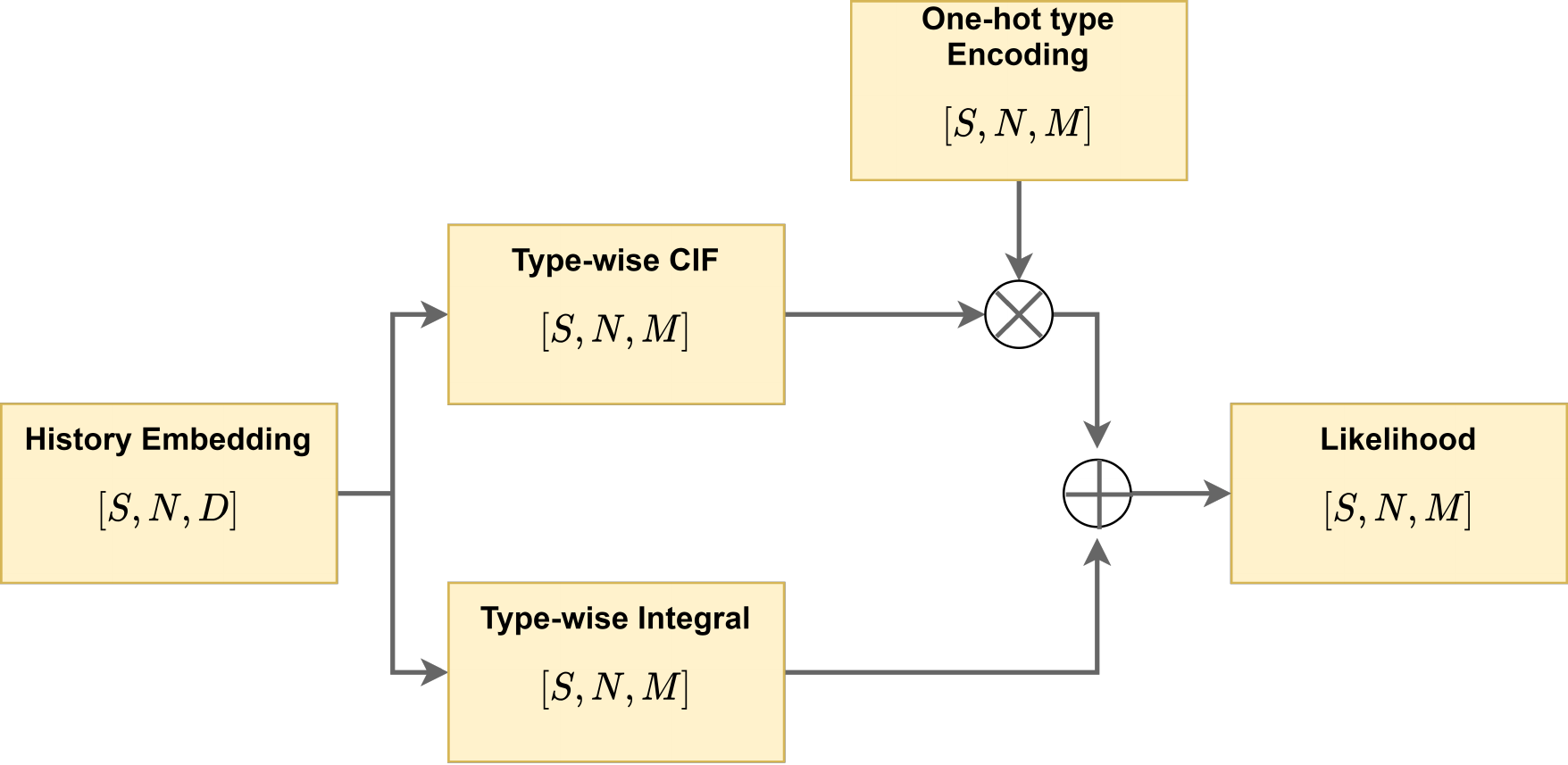}
  \caption{Computation flows of likelihood term, where the notation is the shape of tensor, and $\oplus$ and $\otimes$ are element-wise plus and times operations.}
  \label{figure:likelihoodcompute}
\end{figure}
\section*{F. Proof of Proposition 1.}

\textbf{Proof:}\\
For the CIF of type $m'$ which is $\lambda_{m'}^*(t)$, the overall impacts of changes of historical events of type $m$ on it can be measured by $\sum_j|\partial\lambda_{m'}^*(t)/\partial t_j|$, for $t_j \in \mathcal{H}_m(t)$. When $\bm{A}_{m,m'} = 0$, 
\begin{align}
    \frac{\partial\lambda_{m'}^*(t)}{\partial t_j} 
    &= \frac{\partial\lambda_{m'}^*(t)}{\partial \bm{\Theta}_{m'}(t)} \frac{\partial\bm{\Theta}_{m'}(t)}{\partial t_j}\\\notag
    &= \frac{\partial\lambda_{m'}^*(t)}{\partial \bm{\Theta}_{m'}(t)} \frac{\partial\bm{\chi}_{m'}(\sum_{l=1}^L\rho_l \bm{A}_{m', m_{i-l}}\bm{h}_{m_{i-l},i-l})}{\partial \sum_{l=1}^L\rho_l \bm{A}_{m', m_{i-l}}\bm{h}_{m_{i-l},i-l}} \cdot \\ &\quad \quad \quad \sum_{l=1}^L\rho_l \bm{A}_{m', m_{i-l}} \frac{\partial \bm{h}_{m_{i-l},i-l}}{\partial t_j}\notag
\end{align}
Note that after training, $\bm{A}$ is approximated by $\mathbb{E}_{\bm{A} \sim P(\bm{A}|\bm{X}_{\text{train}})}[\bm{A}]$, which is fixed, so $\frac{\partial \bm{A}_{m', m_{i-1}}}{\partial t_j} = 0$.
For $m_{i-l} \not = m$,  $\bm{h}_{m_{i-l},i-l}$ is computed with the intra-type sequence, which is unrelated to $\mathcal{H}_m(t)$, so $\frac{\partial \bm{h}_{m_{i-l},i-l}}{\partial t_j} = 0$. For $m_{i-l} = m$, so $\bm{A}_{m', m_{i-l}} = 0$. Therefore, for any $t_j \in \mathcal{H}_m(t)$, $\frac{\partial\lambda_{m'}^*(t)}{\partial t_j} = 0$, and all the impacts from historical events are zero. The above claim can be established across the whole time line, and thus the proposition is proved.

\section*{G. Extra details of experiments}
\label{app:G}
\subsection*{G.1. Dataset descriptions}
\label{app:G1}
\textbf{MOOC\footnote{\url{https://github.com/srijankr/jodie/}}.}  The dataset describes the interaction of students with an online course system, with event types are marked on interactions. It is split into 5047, 700, 1300 for training, validation and test sets respectively.
\\
\\
\textbf{Stack Overflow\footnote{\url{https://archive.org/details/stackexchange}}.}  Users of a question-answering website get rewards (called badges) over time for participation, with event types are marked on 22 types of users who are most active in the community. The dataset is split into 4633, 700, 1300 for training, validation and test sets respectively.

\subsection*{G.2. Details of hyperparameters in the reported results}
\label{app:G2}
We give the hyper-parameter settings in Table.~\ref{tab:frameworkresult} for reproduction.
\begin{table}[htb]\resizebox{1.02\columnwidth}{!}{
    \begin{tabular}{l|cccc}
    \toprule
        & Methods              & \multicolumn{1}{c}{learning rate} & layer number & embed dim \\
    \midrule
    \multirow{6}{*}{\rotatebox{90}{MOOC}}         & GRU+LogNormMix       & 0.001         & 2            & 32        \\
                                    & GRU+GomptMix         & 0.01          & 3            & 8         \\
                                    & GRU+WeibMix          & 0.0005        & 2            & 16        \\
                                    & Attention+LogNormMix & 0.001         & 3            & 16        \\
                                    & Attention+GomptMix   & 0.001         & 1            & 16        \\
                                    & Attention+WeibMix    & 0.0005        & 3            & 8         \\
    \midrule
    \multirow{6}{*}{\rotatebox{90}{Stack Overflow}} & GRU+LogNormMix       & 0.0001        & 1            & 8         \\
                                    & GRU+GomptMix         & 0.001         & 2            & 16        \\
                                    & GRU+WeibMix          & 0.0001        & 3            & 32        \\
                                    & Attention+LogNormMix & 0.001         & 1            & 16        \\
                                    & Attention+GomptMix   & 0.0001        & 2            & 32        \\
                                    & Attention+WeibMix    & 0.01          & 3            & 32       
    \\\bottomrule
    \end{tabular}
    
}\vspace*{0.2cm}
\caption{Hyper-parameters used as the best model in Table.~\ref{tab:frameworkresult}.}
    \end{table}
    \\
Besides, the batch size for MOOC is 12, and for Stack Overflow, it is 16. Lag-steps are all set as 32. Mixture components are all set as 16.

\end{document}